\documentclass[10pt,twocolumn,letterpaper]{article}

\usepackage{cvpr}              
\definecolor{cvprblue}{rgb}{0.21,0.49,0.74}
\usepackage[pagebackref,breaklinks,colorlinks,allcolors=cvprblue]{hyperref}

\usepackage{pifont} 
\usepackage{subcaption}
\usepackage{verbatim}
\usepackage{graphicx}
\usepackage{multirow}
\usepackage{xspace}
\usepackage{amsfonts,amssymb} 
\usepackage{microtype}
\usepackage{amsmath}
\usepackage{amsthm}
\usepackage{booktabs}
\usepackage{times}
\usepackage{latexsym}
\usepackage{color}
\usepackage[dvipsnames, table]{xcolor}
\usepackage{soul}
\usepackage{verbatim}
\usepackage{multirow}
\usepackage{xspace}
\usepackage{amsfonts,amssymb} 
\usepackage{microtype}
\usepackage{times}
\usepackage{latexsym}
\usepackage{color}
\usepackage[dvipsnames, table]{xcolor}
\usepackage{soul}
\usepackage{colortbl}
\usepackage{tcolorbox}
\usepackage{booktabs,makecell, multirow, tabularx}
\usepackage{adjustbox}
\usepackage{xcolor}
\usepackage{listings}
\usepackage{amsmath}
\usepackage{gradient-text}
\usepackage{booktabs}
\usepackage{amssymb}
\usepackage{bbding}
\usepackage{pifont}
\usepackage{wasysym}
\usepackage{utfsym}
\usepackage{fontawesome}

\usepackage{mdframed}

\makeatletter

\newcommand{\Rmnum}[1]{\expandafter\@slowromancap\romannumeral #1@}
\makeatother

\newcommand{\modelgradient}{\gradientRGB{UFVideo}{247,97,56}{235,187,12}}

\newcommand{\cmark}{\ding{51}}

\def\paperID{2620} 
\def\confName{CVPR}
\def\confYear{2026}

\title{\modelgradient: Towards Unified Fine-Grained Video Cooperative Understanding with Large Language Models}

\author{
\textbf{Hewen Pan\textsuperscript{1,2}\footnotemark[1]~}, \quad
\textbf{Cong Wei\textsuperscript{2}\footnotemark[1]~}, \quad
\textbf{Dashuang Liang\textsuperscript{2}\footnotemark[2]~}, \quad
\textbf{Zepeng Huang\textsuperscript{2}}, \quad
\textbf{Pengfei Gao\textsuperscript{2}}, \\
\textbf{Ziqi Zhou\textsuperscript{1}}, \quad
\textbf{Lulu Xue\textsuperscript{1}}, \quad
\textbf{Pengfei Yan\textsuperscript{2}}, \quad
\textbf{Xiaoming Wei\textsuperscript{2}}, \quad
\textbf{Minghui Li\textsuperscript{1}}\footnotemark[2]~, \quad
\textbf{Shengshan Hu\textsuperscript{1}}\\
\textbf{$^{1}$}Huazhong University of Science and Technology \quad
\textbf{$^{2}$}Meituan \\
\tt\footnotesize hewenpan@hust.edu.cn~~~weic22@mails.tsinghua.edu.cn~~~liangdas1986@163.com \\
}

\begin{document}

\definecolor{color_red}{RGB}{255,152,150}
\definecolor{color_blue}{RGB}{171,208,241}
\definecolor{color_gray}{RGB}{221,221,221}
\cellcolor{color_blue} \cellcolor{color_pink} \cellcolor{color_yellow} \cellcolor{blue!8}

\twocolumn[{
\renewcommand\twocolumn[1][]{#1}
\maketitle
\begin{center}
    \centering
    \vspace*{-.5cm}
    \includegraphics[width=1.0\textwidth]{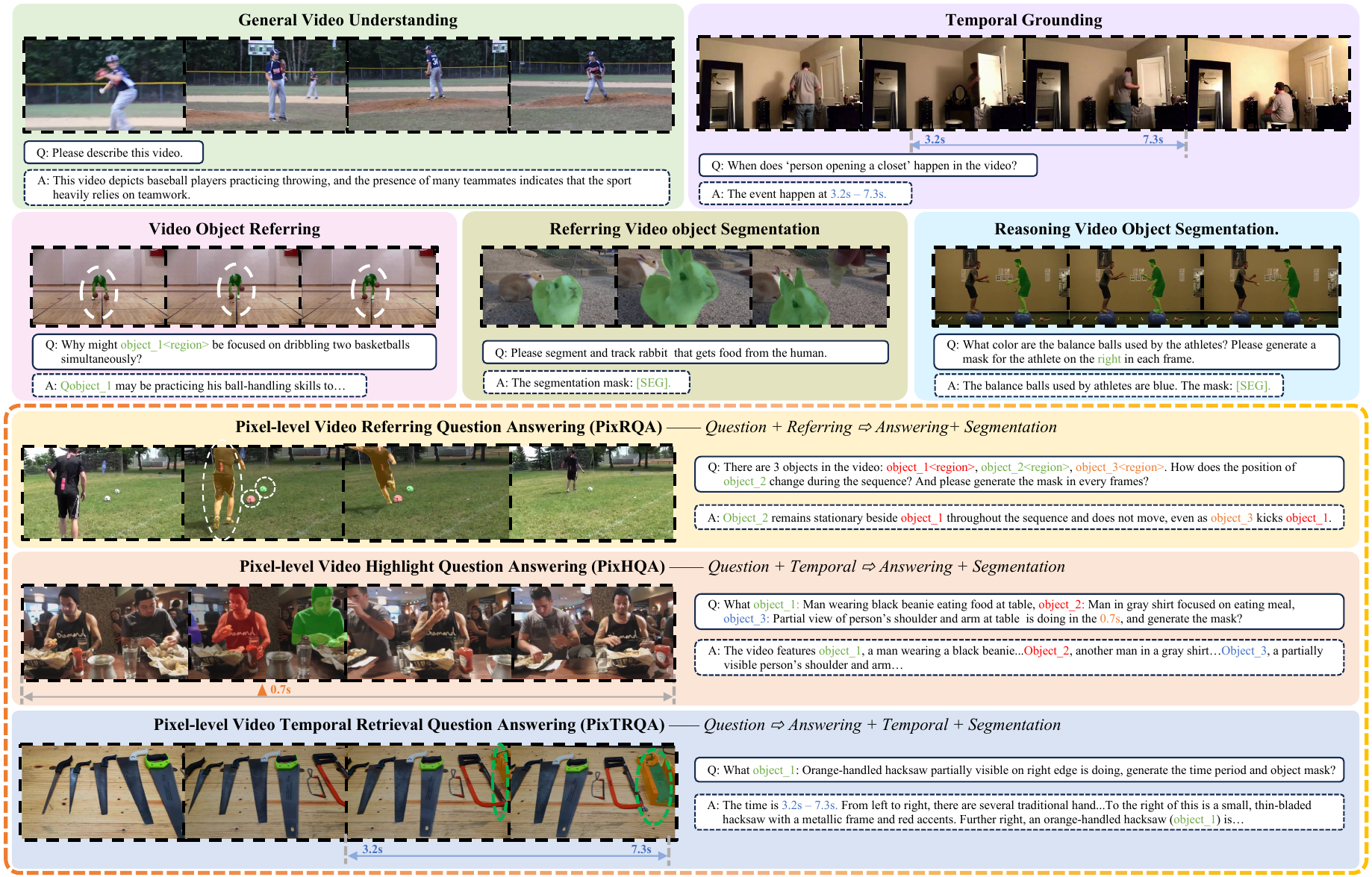}
    \captionof{figure}{Illustration of the video understanding tasks supported by \textbf{\modelgradient}. To the best of our knowledge, UFVideo is the first unified method flexibly supports general video understanding, temporal video grounding, video object referring, referring and reasoning video object segmentation. Representative examples from a novel \textbf{UFVideo-Bench} demonstrating joint multi-grained video cooperative understanding through three tasks: \textbf{PixRQA}, \textbf{PixHQA} and \textbf{PixTRQA}. }
\label{fig:UniMV}
\end{center}
}]

\let\thefootnote\relax\footnotetext{
$^*$ Equal contribution \hspace{5pt}$^\dagger$ Corresponding author
}

\begin{abstract}
With the advancement of multi-modal Large Language Models (LLMs), Video LLMs have been further developed to perform on holistic and specialized video understanding. However, existing works are limited to specialized video understanding tasks, failing to achieve a comprehensive and multi-grained video perception.
To bridge this gap, we introduce \textbf{UFVideo}, the first Video LLM with \textbf{unified multi-grained cooperative understanding} capabilities. Specifically, we design unified visual-language guided alignment to flexibly handle video understanding across global, pixel and temporal scales within a single model. UFVideo dynamically encodes the visual and text inputs of different tasks and generates the textual response, temporal localization, or grounded mask. Additionally, to evaluate challenging multi-grained video understanding tasks, we construct the \textbf{UFVideo-Bench} consisting of three distinct collaborative tasks within the scales, which demonstrates UFVideo's flexibility and advantages over GPT-4o.
Furthermore, we validate the effectiveness of our model across 9 public benchmarks covering various common video understanding tasks, providing valuable insights for future Video LLMs. 
Our code, data and benchmark are available at \url{https://github.com/Heven-Pan/UFVideo}.

\end{abstract}    
\section{Introduction}
\label{sec:intro}

\begin{figure*}[!h]
\centering
\includegraphics[width=1\textwidth]{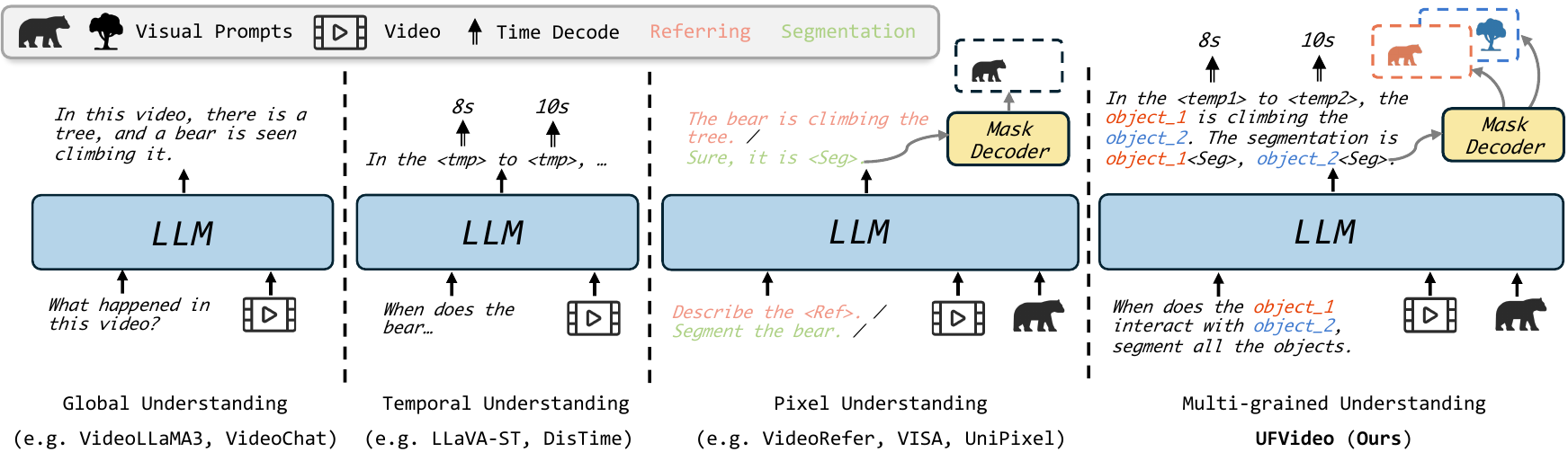}
\caption{Comparison between UFVideo and previous global and fine-grained paradigms, demonstrating UFVideo's ability to flexibly understand video at global, pixel-level and temporal-level granularities.}
\vspace{-0.3cm}
\label{fig:compare_method} 
\end{figure*}

Multi-modal Large Language Models (MLLMs) \cite{hurst2024gpt, bai2025qwen2, qwen3technicalreport, huang2025step, li2023blip} have recently demonstrated strong capabilities in multi-modal understanding by encoding multi-modal information into tokens that align with text tokens in the same space, facilitating comprehension and generation in textual form. Video Large Language Models (Video LLMs) are constructed similarly to perceive videos. 
Initial works on Video LLMs \cite{cheng2024videollama, zhang2025videollama, wang2024qwen2, liu2023visual} leverage large-scale data for alignment training to endow Video LLMs with general video understanding capabilities such captioning \cite{caba2015activitynet, zhou2018towards, patraucean2023perception} and question answering \cite{li2024mvbench, fu2025video, grauman2022ego4d}. To address the growing demand for fine-grained video understanding, subsequent researchers design specialized modules and training strategies to achieve various fine-grained understanding tasks, including video referring \cite{yuan2025videorefer, chen2023shikra, qiu2024artemis, heo2025omni}, video segmentation \cite{wei2024hyperseg, wei2024instructseg, lai2024lisa}, and temporal video grounding \cite{gao2017tall, li2025llava, zeng2025distime}. Specifically, video object referring and video segmentation are categorized under pixel-level video understanding, while temporal video grounding pertains to temporal-level understanding. Notably, these specialized Video LLMs perform tasks independently, failing to effectively integrate fine-grained perception and reasoning for mutual enhancement in understanding \cite{liu2025unipixel, wang2025object}. 

To tackle this challenge, RGA3 \cite{wang2025object} and UniPixel \cite{liu2025unipixel} unify pixel-level video understanding by integrating video object referring and video segmentation, achieving mutual enhancement between perception and reasoning. However, they lack fine-grained temporal understanding. LLaVA-ST \cite{li2025llava} proposes complex spatial-temporal understanding of videos, but is restricted to coarse-grained understanding using bounding boxes. With respect to existing Video LLMs, as illustrated in~\cref{fig:compare_method}, \textbf{these granularities are isolated without explicit correlation during generation}, each targeting specific granularities and lacking generalization across others. For example, models specialized in referring tasks struggle with event timing, while those focused on temporal grounding fail at pixel-level segmentation. However, more importantly, \textbf{video knowledge at different granularities can complement each other}, fine-grained temporal knowledge enhances comprehension of referring objects, and holistic video knowledge can support semantic responses in fine-grained tasks \cite{yuan2025videorefer, li2025llava, zeng2025distime}.

To bridge this gap, we propose \textbf{\modelgradient}, a Video LLM designed to achieve unified fine-grained video understanding tasks, including global video understanding, pixel-level video referring and segmentation, and temporal-level video grounding tasks, enabling a more comprehensive multi-grained understanding of videos. To effectively unify multiple tasks, we design a unified visual-language guided alignment strategy that distinguishes understanding and generation for each task. Visual and textual inputs are uniformly encoded as tokens, respectively. For model generative, textual output and temporal localization are based on LLM, while the SAM2 mask decoder \cite{ravi2024sam} is introduced to map language embeddings to pixel-level segmentation masks. Consequently, UFVideo efficiently performs multi-grained video understanding tasks within a single model. 

Furthermore, to address the limitation in multi-grained video cooperative understanding tasks, we propose a novel multi-grained video understanding benchmark named \textbf{UFVideo-Bench}. Compared to existing datasets, UFVideo-Bench integrates global video understanding, pixel-level video referring and segmentation, and temporal-level video grounding tasks, forming three types of video cooperative understanding tasks. This requires the model to comprehensively understand single or multiple objects within a video at specific time or intervals. UFVideo-Bench facilitates comprehensive multi-grained video understanding. We evaluate the effectiveness of UFVideo on 9 public benchmarks covering various common video understanding tasks, demonstrating UFVideo achieves state-of-the-art performance across those video understanding tasks.
Our contributions are summarized as follows:
\begin{itemize}
    \item We present the UFVideo model, which is the first Video LLM capable of unified fine-grained video cooperative understanding across global, pixel and temporal scales, achieving a common architecture and training strategy across multiple tasks.
    \item To address the lack of a comprehensive benchmark for video cooperative understanding, we are the first to introduce a novel UFVideo-Bench, which integrates general video understanding, video object referring, video segmentation, and temporal grounding tasks.
    \item We assessed the effectiveness of our model across 9 public benchmarks covering various common video understanding tasks and our cooperative understanding benchmarks. As a unified model, it achieved state-of-the-art performance in each of these tasks.
\end{itemize}

\vspace{-0.1cm}
\section{Related Works}

\subsection{Video Large Language Models}

Mainstream Multi-modal Large Language Models (MLLMs) \cite{liu2023visual, huang2025step, hurst2024gpt, qwen3technicalreport, bai2025qwen2} have achieved significant advances in multi-modal understanding by mapping multi-modal data into a unified space for processing by the LLM. This has enabled Video Large Language Models (Video LLMs) \cite{maaz2023video, bai2025qwen2, yuan2025videorefer, lin2025perceive, wei2024instructseg, zeng2025distime, zhang2025videollama} to leverage the powerful capabilities of LLMs for video understanding. To address the growing demands for various video understanding tasks, existing Video LLMs \cite{yuan2025videorefer, lin2025perceive, wei2024instructseg, zeng2025distime} typically require specific architectures to better handle particular understanding tasks. While specialized models can excel at certain tasks, they often face limitations in other understanding scenarios. For example, models design for general video understanding struggle to accurately capture object relationships \cite{yuan2025videorefer}, and those specialized for temporal grounding tasks \cite{zeng2025distime, li2025llava, wang2024grounded} fail to effectively segment objects in videos.

\subsection{Video Understanding tasks with Video LLMs}

Current video understanding tasks can be categorized into three types. Firstly, for global video understanding, Video LLMs need to holistic understanding of videos to perform description and question answering \cite{cheng2024videollama, zhang2025videollama, li2023videochat, maaz2024videogpt+}. Although MLLMs design for images can process video data, Video LLMs require training in video-specific datasets to achieve better performance. 
Pixel-level video understanding encompasses core tasks like video object referring and video segmentation, both aiming to decode fine-grained spatial semantics from visual input but differing in their output objectives. For video object referring, Video LLMs are designed to comprehend semantic information of specific regions in videos, enabling region description and question answering. Some MLLMs for images \cite{zhang2024gpt4roi,wang2024cogvlm, peng2023kosmos, chen2023shikra} enhance region comprehension by incorporating visual prompts, such as bounding boxes or pixel-level masks. However, in pixel-level video understanding, video object referring faces more complex challenges from temporal aspects and continuous region tracking, prompting recent works \cite{yuan2025videorefer, lin2025perceive, lian2025describe, sun2025sama} to design video-specific visual prompt modules. 
For the video segmentation task, it emphasizes Video LLMs can provide pixel-level masks for corresponding objects. Previous researchs on MLLMs \cite{lai2024lisa, pi2024perceptiongpt, ren2024pixellm, xia2024gsva} focus on segmentation within the image modality, typically generating segmentation tokens for static images using MLLM and then employing decoders such as SAM to produce masks. In video modality, the impact of temporal dynamics must be considered. Consequently, some works \cite{wei2024instructseg, yan2024visa, wei2024hyperseg} are designed to address multi-frame sampling and tracking of objects across frames.
Lastly, the temporal-level video understanding task is enable Video LLMs to understand the occurrence and relationship of requested events, and locate precise time to showcase comprehension of these events. For instance, Dense Video Captioning \cite{huang2020multimodal, tang2019coin} requires detailing all events in the video along with their timeline, moment retrieval \cite{gao2017tall, grauman2022ego4d} necessitates identifying the time of requested events within the video, and grounded video question answering \cite{huang2024vtimellm, xiao2024can} involves understanding the video based on given temporal.

\section{Method}
\label{sec:model}

\begin{figure*}[!h]
\centering
\includegraphics[width=0.9\textwidth]{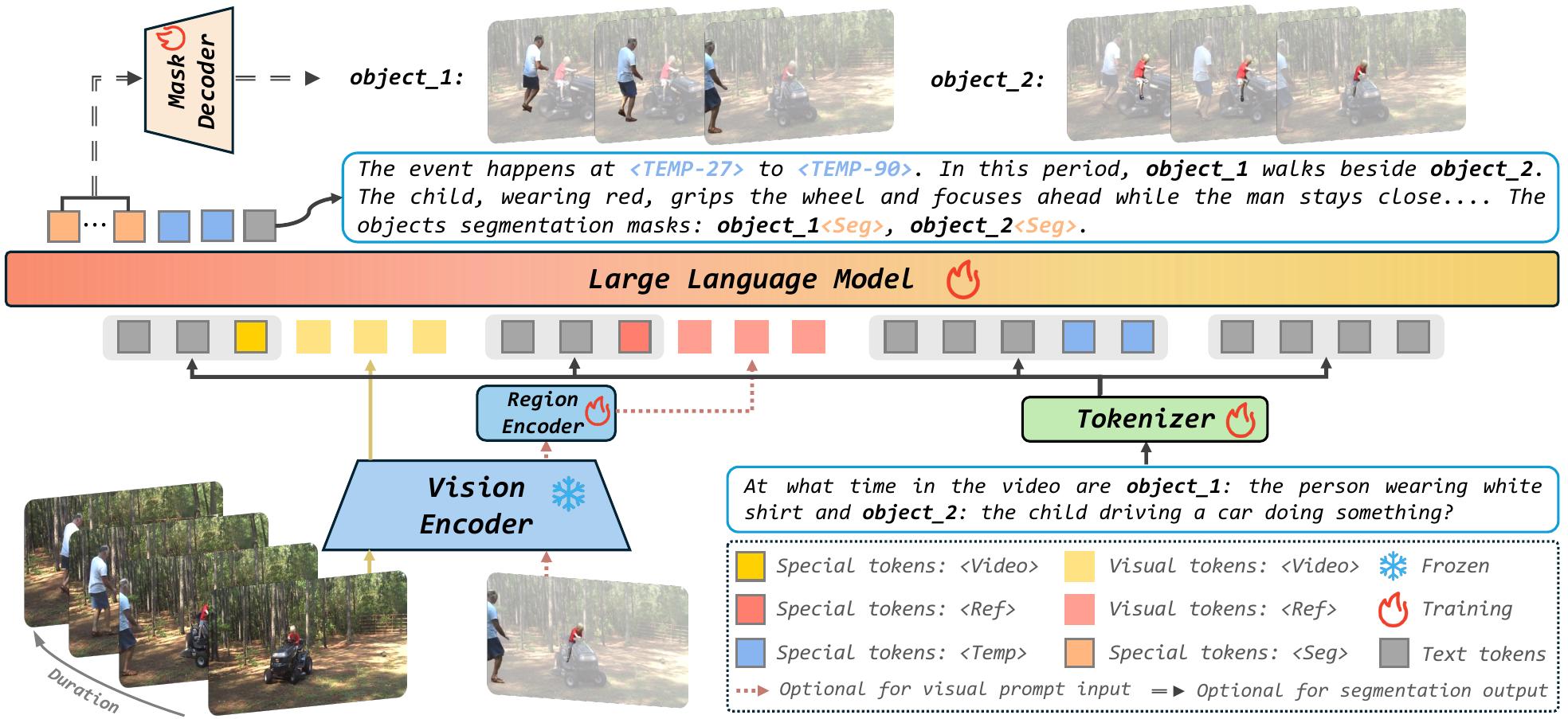}
\caption{Overall Architecture of UFVideo.}
\vspace{-0.3cm}
\label{fig:pipeline} 
\end{figure*}

\paragraph{Problem Formulation.}
We provide a unified definition for the multi-grained video understanding task. Given a video $V \in \mathbb{R}^{N\times C\times W\times H}$, where $N$, $C$, $W$, $H$ represent its frame numbers, channels, width, and height, respectively, and a question prompt $Q$ and object visual prompts $M \in \mathbb{R}^{N_s \times K_I\times W\times H}$ (optional), where $M$ are the masks of $N_s$ objects in $Q$ in input-specific $K_I$ frames. Our goal is to enable video LLMs to output the textual answering, including language generation and temporal localization (optional), and segmentation masks for objects in $Q$ (optional).
\begin{equation}
    A, S, T = \Phi \left( Q, V, M \right)
\end{equation}
Here, $\Phi$ denotes the Video LLMs, $A$ denotes the language response, $T$ represents the time located in the video, and $S \in \mathbb{R}^{N_s \times K_O\times W\times H}$ represents the masks obtained for all $N_s$ objects during the subsequent frames of the event $K_O$.

\vspace{-0.3cm}
\paragraph{Overview.}
Our model architecture is depicted in~\cref{fig:pipeline}. We introduce a Video LLM that achieves unified fine-grained video understanding across global, pixel-level and temporal-level. Following previous works, our model employs an LLM as the backbone and uses a vision encoder to convert video inputs into discrete tokens, which are then aligned with text tokens. To unify multiple tasks, we design unified alignment training to distinguish understanding and generation for each task. Additionally, we incorporate the SAM2 mask decoder \cite{ravi2024sam} into the training process to generate segmentation masks, facilitating pixel-level segmentation.

\vspace{-0.05cm}

\subsection{Multi-Grained Video Tasks Alignment}
\label{sec:multi-grained_alignment}

\paragraph{Motivation.}
Video LLMs have been developed to interpret videos for general question answering and various fine-grained understanding tasks. However, as illustrated in~\cref{fig:compare_method}, prior works often concentrate on isolated granularity paradigms, which hard to meet the complex demands of real-world. Existing works \cite{liu2025unipixel, li2025llava, wang2025object} demonstrate that integrating global and fine-grained details across tasks can enhance the comprehensive understanding of Video LLMs. Therefore, our work aims to improve the comprehensive perception capabilities of Video LLMs, enabling them to flexibly handle video cooperative understanding tasks. Inspired by the successful fine-grained task paradigm \cite{yuan2025videorefer, yan2024visa, li2025llava}, we strive to unify global, pixel-level and temporal-level understanding tasks within a single model.

\vspace{-0.3cm}

\paragraph{Visual-Language guided Alignment.}
\label{para:alignment}
To effectively distinguish and unify various video understanding tasks, we design special tokens for language prompts to align multi-grained video understanding.
For textual instructions, we follow previous work by adopting next-token generation, and these instruction are tokenized into textual tokens $\mathcal{T}_i \in \mathbb{R}^{L_V \times P}$, which $L_V$ represents the vocabulary size of the tokenizer and $P$ denotes the hidden size of projection layers.
To accommodate different video durations, we design relative temporal tokens \texttt{<Temp-$\tau$>}, which is used to represent time in both input and output of the model. All video duration are sampled to a fixed length denoted as $N_t$.
\begin{equation}
    \tau = \frac{t}{T_n} \times N_t
\end{equation}
where $t$ is the time of the current frame and $T_n$ denotes the total duration of the video. The temporal tokens are tokenized into $\mathcal{T}_t \in \mathbb{R}^{N_t \times P}$.
For video object referring task, we insert the special token \texttt{<Ref>} as a placeholder, which is only used for the model input. Through the \texttt{<Ref>} tokens, the corresponding object visual prompts $M$ are injected into the LLM input.
In addition, we insert special tokens \texttt{<Seg>} for video segmentation tasks. To generate the segmentation mask from the LLM output language embedding, we position the masks of \texttt{<Seg>} tokens to extract the segmentation related language embedding.
\begin{equation}
    \rho_{s} = \mathbb{I}\left(\mathcal{T}_i, \, \mathcal{T}_s\right)
\end{equation}
where $\mathcal{T}_s$ represent the tokenized \texttt{<Seg>} tokens. $\mathbb{I}$ represents the indicator function of locating the $\mathcal{T}_s$ in the $\mathcal{T}_i$, $\rho_s$ denotes the position mask of \texttt{<Seg>}.

Therefore, we ulitize the \texttt{<Temp-$\tau$>}, \texttt{<Ref>} and \texttt{<Seg>} tokens to achieve multi-grained coorperative video understanding within a single model.

\subsection{UFVideo Architecture}
\paragraph{Encode for Multi-Modal Input.}
\label{para:encode}

For $V$ and its corresponding object visual prompt $M$ (optional) across all tasks, since $M$ contains detailed information about objects $O$ required to be referred in the video referring task, we employ a pre-trained vision encoder $\Phi_{v}$ to separately encode $V$ and $M$, obtaining video features $F_V \in \mathbb{R}^{N\times W\times H \times D}$ and object visual prompt features $F_M \in \mathbb{R}^{K_I \times W\times H \times D}$, where $D$ represents the dimensions of visual features. And then we apply methods from VideoRefer \cite{yuan2025videorefer} to get object visual prompt spatial features $S_M \in \mathbb{R}^{K_I\times D}$ from $F_M$. Therefore, $S_M$ is projected into object visual tokens $\mathcal{T}_r \in \mathbb{R}^{U \times P}$, which are injected into the LLM input following the \texttt{<Ref>} token setting in~\cref{para:alignment}, where $U$ is the merge size after compressing $S_M$. 
For the video segmentation task, we randomly select $K$ frames $V_{s}$ from the video and encode them using the Hiera-L encoder denoted as $\mathbf{E}$ from SAM2, which serves as the visual input to train the mask decoder denoted as $\mathbf{D}$.

\vspace{-0.5cm}
\paragraph{Decode for LLM Generative.}
Given that we integrate different tasks, the LLM is expected to generate results required by various tasks, such as language generation, precise time, and pixel-level spatial segmentation.  
\begin{equation}
    \mathrm{H} = \mathbf{U}_p\left( \mathcal{T}_v,\, \mathcal{T}_i,\, \mathcal{T}_r,\, \mathcal{T}_t^{I} \right)
\end{equation}
where $\mathbf{U}_p$ represent the LLM,  $\mathrm{H}$ is the last hidden state projected by the $\mathbf{U}_p$, and $\mathcal{T}_t^{I}$ denotes the input form of temporal token $\mathcal{T}_t$.
For textual language generation and precise time, they can both be generated using text-form tokens.
\begin{equation}
    \mathcal{Y}_t = p_{\theta} \left( \mathrm{H} \right) \quad
    \mathcal{Y}_m = p_{\theta}\left( \mathrm{H} \right) \times \frac{T_n}{N_t}
\end{equation}
where $\mathcal{Y}_t$ and $\mathcal{Y}_m$ are the textual language and the output timing, $p_{\theta}$ represent the projection layer for LLM output embedding $\mathrm{H}$.
However, for pixel-level segmentation, it is challenging for the LLM to directly output tokens corresponding to segmentation masks. Therefore, we incorporate SAM2's mask decoder and introduce a special token \texttt{<Seg>} in ~\cref{sec:multi-grained_alignment} for the segmentation task. We extract the segmentation embeddings from $\mathrm{H}$ to train the mask decoder together with \( V_{s} \). Due to the varying number of objects requested for segmentation in different samples, dynamic embedding training is required for each batch.
\begin{equation}    
    \mathrm{\textbf{S}} = \mathbf{D}\left( \sum_{i=1}^{B} \, \left(\mathbf{E}\left(V_{s(i)}\right), \theta\left(\mathrm{H}_{(i)}\right)\odot \rho_{s(i)}^{\top} \right) \right)
\end{equation}
where $\theta$ is the projection of the fully connected layer, $B$ is the size of the training batch, position mask $\rho_s$ performs element-wise multiplication $\odot$ with the projected language embedding to produce the segmentation-specific language embeddings. $\mathrm{\textbf{S}} = \{ S_{(1)}, S_{(2)}, \ldots, S_{(B)}\}$ denotes the batch mask output.

\vspace{-0.4cm}
\paragraph{Training Objective.}

For our proposed unified video LLMs, we convert the visual and textual inputs from multiple tasks into discrete tokens. This allows us to train our LLM using the common next-token prediction objective of language modeling.
\begin{equation}
    \mathcal{L}_{text} = - \frac{1}{N} \sum_{i=1}^{N} \log p_{\theta} \left( x_{i+1} | x_1,x_2,\ldots,x_i \right)
\end{equation}
Here, $\{x_1, x_2, \ldots, x_n\}$ represents the unified token sequence, $N$ is the length of the token sequence, and $\log p_{\theta}$ is the negative log-likelihood function of the LLM used to predict the logarithmic probability of the target token.

Additionally, to train mask decoder to learn the mapping from the LLM-generated \texttt{<Seg>} token to masks, we employ binary cross-entropy loss (BCE) and DICE loss to constrain the training process based on the ground-truth mask $S_t$ and the prediction mask $S_p$.
\begin{equation}
    \begin{split}
        \mathcal{L}_{mask} = \alpha \cdot \textit{BCE}\left( S_p, S_t \right) + \beta \cdot \textit{DICE} \left(  S_p, S_t \right)
    \end{split}
\end{equation}
Here, $\alpha$ and $\beta$ are the learning rate weights for BCE loss and Dice loss, respectively. Therefore, we set the weight of the $\mathcal{L}_{text}$ is $\gamma$, the overall training objective for the model can be formulated as:
\begin{equation}
    \mathcal{L} = \gamma \cdot \mathcal{L}_{text} + \mathcal{L}_{mask}
\end{equation}

\vspace{-0.4cm}

\paragraph{Difference between UFVideo and Previous Work.}

Compared with previous Video LLMs \cite{yuan2025videorefer, zeng2025distime, wei2024hyperseg, cheng2024videollama}, UFVideo is the first model to unify the task form of multi-grained video understanding, including global video question-answering, pixel-level video understanding and temporal-level video understanding tasks, enjoying the benefits of multi-grained and multi-task joint video learning as described in~\cref{ablation_study}. In addition, previous tasks are separate, but real-world scenarios are complex and composite, requiring the collaboration of multiple fine-grained tasks. Thus, we integrate the above video understanding tasks to construct UFVideo-Bench.

\begin{table*}[t]
\setlength{\tabcolsep}{3pt} 
\centering
\caption{Exprimental results on MVBench \cite{li2024mvbench} for general video understanding tasks, and comparison with state-of-the-art methods. The \textbf{bold} and \underline{underlined} text indicate the best and second-best results.}
\renewcommand{\arraystretch}{1}
\adjustbox{width=1\linewidth}{
   \begin{tabular}{l|c|cccccccccccccccccccc|c}
   
       \toprule[1.2pt]
            \textbf{Model} & \textbf{Size} & \textbf{AS} & \textbf{AP} & \textbf{AA} & \textbf{FA} & \textbf{UA} & \textbf{OE} & \textbf{OI} & \textbf{OS} & \textbf{MD} & \textbf{AL} & \textbf{ST} & \textbf{AC} & \textbf{MC} & \textbf{MA} & \textbf{SC} & \textbf{FP} & \textbf{CO} & \textbf{EN} & \textbf{ER} & \textbf{CI} & \textbf{Avg. $\uparrow$} \\
\midrule
GPT-4V \cite{openai2023gpt4v} & -- & 55.5 & 63.5  & 72.0 & 46.5 & 73.5 & 18.5 & 59.0 & 29.5 & 12.0 & 40.5 & 83.5 & 39.0 & 12.0 & 22.5 & 45.0 & 47.5 & 52.0 & 31.0 & 59.0 & 11.0 & 43.5 \\
VideoLLaMA2.1 \cite{cheng2024videollama} & 7B & -- & -- & -- & -- & -- & -- & -- & -- & -- & -- & -- & -- & -- & -- & -- & -- & -- & -- & -- & -- & 57.3 \\
RGA3 \cite{wang2025object} & 7B & -- & -- & -- & -- & -- & -- & -- & -- & -- & -- & -- & -- & -- & -- & -- & -- & -- & -- & -- & -- & 63.8 \\
Qwen2-VL \cite{wang2024qwen2} & 7B & -- & -- & -- & -- & -- & -- & -- & -- & -- & -- & -- & -- & -- & -- & -- & -- & -- & -- & -- & -- & \underline{67.0} \\
TimeChat \cite{ren2024timechat} & 7B & 40.5 & 36.0 & 61.0 & 32.5 & 53.0 & 53.5 & 41.5 & 29.0 & 19.5 & 26.5 & 66.5 & 34.0 & 20.0 & 43.5 & 42.0 & 36.5 & 36.0 & 29.0 & 35.0 & 35.0 & 38.5 \\
Video-LLaVA \cite{lin2024video}  & 7B & 46.0 & 42.5 & 56.5 & 39.0 & 53.5 & 53.0 & 48.0 & 41.0 & 29.0 & 31.5 & 82.5 & 45.0 & 26.0 & 53.0 & 41.5 & 33.5 & 41.5 & 27.5 & 38.5 & 31.5 & 43.0 \\
PLLaVA \cite{xu2024pllava} & 7B & 58.0 & 49.0 & 55.5 & 41.0 & 61.0 & 56.0 & 61.0 & 36.0 & 23.5 & 26.0 & 82.0 & 39.5 & 42.0 & 52.0 & 45.0 & 42.0 & 53.5 & 30.5 & 48.0 & 31.0 & 46.6 \\
VideoGPT+ \cite{maaz2024videogpt+} & 4B & 69.0 & 60.0 & 83.0 & 48.5 & 66.5 & 85.5 & 75.5 & 36.0 & 44.0 & 34.0 & 89.5 & 39.5 & 71.0 & 90.5 & 45.0 & 53.0 & 50.0 & 29.5 & 44.0 & 60.0 & 58.7 \\
VideoChat2 \cite{li2024mvbench} & 7B & 75.5 & 58.0 & 83.5 & \underline{50.5} & 60.5 & 87.5 & 74.5 & \textbf{45.0} & 47.5 & 44.0 & 82.5 & 37.0 & 64.5 & 87.5 & \textbf{51.0} & \textbf{66.5} & 47.0 & 35.0 & 37.0 & 72.5 & 60.4 \\
UniPixel \cite{liu2025unipixel} & 3B & 69.5 & \underline{62.5} & 83.0 & 48.5 & \textbf{76.5} & 86.5 & 66.5 & 38.0 & \textbf{49.0} & 40.5 & 87.0 & \textbf{49.0} & 74.0 & 95.0 & \underline{49.0} & 45.0 & \underline{63.5} & 34.5 & \textbf{58.0} & \underline{73.5} & 62.5 \\
LLaVA-ST \cite{li2025llava} & 7B & \underline{77.0} & \textbf{69.0} & \underline{91.5} & 50.0 & 68.5 & \textbf{93.5} & \underline{84.5} & 40.0 & 44.5 & \underline{49.5} & \underline{93.0} & \underline{44.0} & \underline{77.5} & \textbf{97.5} & 41.0 & 57.0 & 56.6 & \textbf{37.0} & 49.0 & 63.0 & 64.2 \\
\midrule[0.5pt]
\textbf{\modelgradient} & 7B & \textbf{81.5} & 61.0 & \textbf{95.0} & \textbf{54.0} & \underline{74.5} & \underline{92.5} & \textbf{89.0} & \underline{43.0} & \underline{48.5} & \textbf{56.6} & \textbf{93.5} & 42.5 & \textbf{81.0} & \underline{96.5} & 47.5 & \underline{59.0} & \textbf{67.0} & \underline{36.5} & \underline{51.0} & \textbf{76.0} & \textbf{67.3}  \\
            \bottomrule[1.2pt]
       \end{tabular}
   }
   \vspace{-0.3cm}
   \label{tab:mvbench}
\end{table*}

\subsection{UFVideo Dataset and Benchmark}

In this section, we introduce the \textbf{UFVideo-Dataset} and \textbf{UFVideo-Bench}, which serve as the training set and benchmark for multi-grained video understanding tasks, respectively. We propose three tasks that integrate global, pixel-level and temporal-level video understanding tasks, involving both single and multiple objects. ~\cref{fig:UniMV} illustrates the typical example in three tasks. (Please refer to the Supplementary Materials for complete details.)
Here are some key details of the three tasks:
\vspace{-0.3cm}
\paragraph{Pixel-level Video Referring Question Answering (PixRQA).} This task joints the global and pixel-level video understanding, which need model outputs the textual response and pixel-level masks based on the textual question, video and single-frame mask prompt input. Additionally, we construct the question according to prompt template 1.
\vspace{-0.4cm}
\paragraph{Pixel-level Video Highlight Question Answering (PixHQA).} This task introduces temporal video grounding compared to PixRQA. The model interprets the video based on time-specific questions, outputting textual answers and segmentation masks for objects at the specific time. We use the Qwen3-VL-235B-A22B-Instruct \cite{qwen3technicalreport} to generate the question and answer for relevant objects in frames at specific times. The construction is based on prompt template 2.
\vspace{-0.3cm}
\paragraph{Pixel-level Video Temporal Retrieval Question Answering (PixTRQA).} This task introduces an additional level of difficulty by transforming the time-grounded video understanding in PixHQA into video moment retrieval. The model must respond to the textual question by performing language generation, temporal localization, and pixel-level segmentation within the video. We maintain consistency with the annotation approach from PixHQA. The question is restructured according to prompt template 3.

\section{Experiment}
\subsection{Implementation Details}
We use SigLIP-so400m-patch14-384 \cite{zhai2023sigmoid} as the vision encoder. The pre-trained model is VideoRefer 7B \cite{yuan2025videorefer}. The $N_t$ is 100, and $ N_s $ is 4. $\alpha$, $\beta$, and $\gamma$ are set to 2.0, 0.5, and 1.0, respectively. The training global batch size is 512 for stage 1 and 256 for stage 2. Stage 1 is trained for 2 epochs, and stage 2 is 1 epoch. We train UFVideo using 32 A800 (80G) GPUs. The DeepSpeed engine is used to efficiently manage distributed training and optimization. The optimizer is set to AdamW, with a WarmupCosineLR learning rate scheduler.

\begin{table}[!htb]
\setlength{\tabcolsep}{2.3pt} 
\centering
\caption{Experimental results on VideoRefer-Bench-D benchmark \cite{yuan2025videorefer} for video referring description task, and comparison with state-of-the-art methods. The \textbf{bold} and \underline{underlined} text indicate the best and second-best results.}
\renewcommand{\arraystretch}{1}
\adjustbox{width=1\linewidth}{
   \begin{tabular}{l|c|ccccc|ccccc}
       \toprule[1.2pt]
            \multirow{2}{*}{Model}
            &\multirow{2}{*}{Size}
            &\multicolumn{5}{c|}{Single-Frame}
            &\multicolumn{5}{c}{Multi-Frame}\\
\cmidrule{3-7} \cmidrule{8-12} 
                & &SC $\uparrow$ &AD $\uparrow$ &TD $\uparrow$ &HD $\uparrow$
                &Avg. $\uparrow$ &SC $\uparrow$ &AD $\uparrow$ &TD $\uparrow$ &HD $\uparrow$
                &Avg. $\uparrow$\\
\hline
\multicolumn{12}{c}{\cellcolor{color_gray}\textit{General MLLMs}} \\
Qwen2-VL\cite{wang2024qwen2}&7B &2.97 & 2.24 & 2.03 & 2.31 & 2.39 & 3.30 & 2.54 & 2.22 & 2.12 & 2.55 \\
InternVL2\cite{team2024internvl2}&26B & 3.55 & 2.99 & 2.57 & 2.25 & 2.84 & 4.08 & 3.35 & 3.08 & 2.28 & 3.20 \\
GPT-4o\cite{hurst2024gpt}& -- & 3.34 & 2.96 & 3.01 & 2.50 & 2.95 & 4.15 & 3.31 & 3.11 & 2.43 & 3.25 \\
\hline
\multicolumn{12}{c}{\cellcolor{color_gray}\textit{Image LLMs}} \\
Ferret\cite{you2023ferret}&7B & 3.08 & 2.01 & 1.54 & 2.14 & 2.19 & 3.20 & 2.38 & 1.97 & 1.38 & 2.23 \\
Osprey\cite{yuan2024osprey}&7B & 3.19 & 2.16 & 1.54 & 2.45 & 2.34 & 3.30 & 2.66 & 2.10 & 1.58 & 2.41 \\
\hline 
\multicolumn{12}{c}{\cellcolor{color_gray}\textit{Video LLMs}} \\
Artemis\cite{qiu2024artemis}&7B & -- & -- & -- & -- & -- & 3.42 & 1.34 & 1.39 & 2.90 & 2.26 \\
PAM\cite{lin2025perceive}& 3B & -- & -- & -- & -- & -- & 3.92 & 2.84 & 2.88 & 2.94 & 3.14 \\
DAM\cite{lian2025describe}& 8B & -- & -- & -- & -- & -- & \textbf{4.69} & \textbf{3.61} & \underline{3.34} & \underline{3.09} & \textbf{3.68} \\
VideoRefer\cite{yuan2025videorefer}&7B & 4.41 & 3.27 & 3.03 & 2.97 & 3.42 & 4.44 & 3.27 & 3.10 & 3.04 & 3.46 \\
UniPixel\cite{liu2025unipixel}&7B & \underline{4.45} & \underline{3.32} & \underline{3.05} & \underline{3.04} & \underline{3.47} & 4.48 & 3.34 & 3.03 & 3.07 & 3.48 \\

\midrule[0.5pt]
\textbf{\modelgradient}& 7B & \textbf{4.53} & \textbf{3.48} & \textbf{3.26} & \textbf{3.09} & \textbf{3.59} & \underline{4.56} & \underline{3.40} & \textbf{3.35} & \textbf{3.15} & \underline{3.61}  \\
            \bottomrule[1.2pt]
       \end{tabular}
   }
   \vspace{-0.3cm}
   \label{tab:videorefer-d}
\end{table}

\begin{table}[htb]
\setlength{\tabcolsep}{4pt} 
\centering
\caption{Experimental results on VideoRefer-Bench-Q benchmark \citet{yuan2025videorefer} for video object referring question answering task, and comparison with state-of-the-art methods. Mode denotes evaluation using single or multiple frames. The \textbf{bold} and \underline{underlined} text indicate the best and second-best results.}
\renewcommand{\arraystretch}{1}
\adjustbox{width=1\linewidth}{
   \begin{tabular}{l|cc|cccccc}
       \toprule[1.2pt]
            \multirow{1}{*}{Model}
            &\multirow{1}{*}{Size}
            &\multirow{1}{*}{Mode}
            &\multicolumn{1}{c}{BQ $\uparrow$}
            &\multicolumn{1}{c}{SQ $\uparrow$}
            &\multicolumn{1}{c}{RQ $\uparrow$}
            &\multicolumn{1}{c}{CQ $\uparrow$}
            &\multicolumn{1}{c}{FP $\uparrow$}
            &\multicolumn{1}{c}{Avg. $\uparrow$}
            \\

\hline
\multicolumn{9}{c}{\cellcolor{color_gray}\textit{General MLLMs}} \\
LLaVA-OV\cite{li2024llava}&7B &S & 58.7 & 62.9 & \textbf{64.7} & 87.4 & 76.3 & 67.4  \\
Qwen2-VL\cite{wang2024qwen2}&7B &S & 62.0 & 69.6 & 54.9 & 87.3 & 74.6 & 66.0 \\
InternVL2\cite{team2024internvl2}&26B &S & 58.5 & 63.5 & 53.4 & 88.0 & 78.9 & 65.0 \\
GPT-4o-mini\cite{hurst2024gpt}& -- &S & 57.6 & 67.1 & 56.5 & 85.9 & 75.4 & 65.8\\
GPT-4o\cite{hurst2024gpt}& -- &S & 62.3 & \textbf{74.5} & 66.0 & 88.0 & 73.7 & 71.3 \\
\hline
\multicolumn{9}{c}{\cellcolor{color_gray}\textit{Image LLMs}} \\
Ferret\cite{you2023ferret}&7B &S & 35.2 & 44.7 & 41.9 & 70.4 & 74.6 & 48.8 \\
Osprey\cite{yuan2024osprey}&7B &S & 45.9 & 47.1 & 30.0 & 48.6 & 23.7 & 39.9 \\
\hline 
\multicolumn{9}{c}{\cellcolor{color_gray}\textit{Video LLMs}} \\
VideoRefer\cite{yuan2025videorefer}&7B &S & 75.4 & 68.6 & 59.3 & \textbf{89.4} & 78.1 & 71.9 \\
UniPixel\cite{liu2025unipixel}&7B &S & 71.7 & 73.2 & \underline{64.6} & 90.1 & \underline{79.6} & 73.8 \\
RGA3\cite{wang2025object}&7B &S & \underline{77.4} & 71.9 & 61.1 & \underline{88.8} & \textbf{81.6} & \underline{74.0} \\
\textbf{\modelgradient} &7B &S & \textbf{80.9} & \underline{73.3} & 62.7 & 86.7 & 78.9 & \textbf{74.9} \\
\midrule[0.5pt]
VideoRefer\cite{yuan2025videorefer}&7B &M & -- & 70.6 & 60.5 & -- & -- & 72.1 \\

UniPixel\cite{liu2025unipixel}&7B &M & \underline{79.5} & \underline{74.7} & \underline{64.4} & \underline{90.8} & \underline{81.5} & \underline{76.3} \\
\textbf{\modelgradient} &7B &M & \textbf{83.8} & \textbf{75.8} & \textbf{64.7} & \textbf{91.6} & \textbf{82.5} & \textbf{77.9} \\
            \bottomrule[1.2pt]
       \end{tabular}
   }
   \vspace{-0.7cm}
   \label{tab:videorefer-q}
\end{table}

\begin{table*}[t]
\setlength{\tabcolsep}{7pt} 
\centering
\caption{Experimental results on MeViS \cite{ding2023mevis} \texttt{val$^\texttt{u}$} , MeViS \cite{ding2023mevis} \texttt{val} , Ref-YouTube-VOS \cite{seo2020urvos} \texttt{val} , Ref-DAVIS17 \cite{pont20172017} \texttt{val} and ReVOS \cite{yan2024visa} \texttt{val} datasets for referring video segmentation and reasoning video segmentation tasks. We present the overall performance of ReVOS, which the referring and reasoning metrics are shown in Appendix. The \textbf{bold} and \underline{underlined} text indicate the best and second-best results.}
\renewcommand{\arraystretch}{1}
\adjustbox{width=1\linewidth}{
   \begin{tabular}{l|c|ccc|ccc|ccc|ccc|ccc}
       \toprule[1.2pt]
            \multirow{2}{*}{Model}
            &\multirow{2}{*}{Size}
            &\multicolumn{3}{c|}{MeViS \texttt{val$^\texttt{u}$}}
            &\multicolumn{3}{c|}{MeViS \texttt{val}}
            &\multicolumn{3}{c|}{Ref-YouTube-VOS \texttt{val}}
            &\multicolumn{3}{c|}{Ref-DAVIS17 \texttt{val}}
            &\multicolumn{3}{c}{ReVOS \texttt{val} (Overall)}\\
\cmidrule{3-5} \cmidrule{6-8} \cmidrule{9-11} \cmidrule{11-14} 
\cmidrule{15-17} 
                & & $\mathcal{J}$ & $\mathcal{F}$ & $\mathcal{J\&F}$
                & $\mathcal{J}$ & $\mathcal{F}$ & $\mathcal{J\&F}$
                & $\mathcal{J}$ & $\mathcal{F}$ & $\mathcal{J\&F}$
                & $\mathcal{J}$ & $\mathcal{F}$ & $\mathcal{J\&F}$
                & $\mathcal{J}$ & $\mathcal{F}$ & $\mathcal{J\&F}$ \\
\hline
\multicolumn{17}{c}{\cellcolor{color_gray}\textit{Specialist Models}} \\
ReferFormer \cite{wu2022language}  & -- & -- & -- & -- & 29.8 & 32.2 & 31.0 & 61.3 & 64.6 & 62.9 & 58.1 & 64.1 & 61.1 & 26.2 & 29.9 & 28.1  \\
LMPM \cite{ding2023mevis}  & -- & 36.5 & 43.9 & 40.2 & 34.2 & 40.2 & 37.2 & -- & -- & -- & -- & -- & -- & 21.2 & 31.7 & 26.4 \\
OnlineRefer \cite{wu2023onlinerefer}  & -- & -- & -- & -- & -- & -- & -- & 61.6 & 65.5 & 63.5 & 61.6 & 67.7 & 64.8 & -- & -- & -- \\
\hline
\multicolumn{17}{c}{\cellcolor{color_gray}\textit{Video LLMs}} \\
PixelLM \cite{ren2024pixellm} & 7B & -- & -- & --  & 36.3 & 41.1 & 38.7 & 54.3 & 55.7 & 55.0 & 63.4 & 70.0 & 66.7 & -- & -- & -- \\
LISA \textcolor{color_gray}{+ XMem} \cite{lai2024lisa} & 13B & 41.9 & 49.3 & 45.6  & 35.8 & 40.0 & 37.9 & 54.0 & 54.8 & 54.4 & 63.2 & 68.8 & 66.0 & 39.1 & 42.7 & 40.9 \\
VISA \cite{yan2024visa} & 13B & -- & -- & -- & 41.8 & 47.1 & 44.5 & 61.4 & 64.7 & 63.0 & 67.0 & 73.8 & 70.4 & 45.3 & 49.7 & 47.5 \\
VideoLISA \textcolor{color_gray}{+ Post} \cite{bai2024one} & 3.8B & 50.9 & 58.1 & 54.5 & 41.3 & 47.6 & 44.4 & 61.7 & 65.7 & 63.7 & 64.9 & 72.7 & 68.8 & -- & -- & -- \\
VideoGLaMM \cite{munasinghe2025videoglamm} & 3.8B & -- & -- & --  & 42.1 & 48.2 & 45.2 & 65.4 & 68.2 & 66.8 & 73.3 & 65.6 & 69.5  & -- & -- & -- \\
ViLLa \cite{zheng2025villa} & 6B & -- & -- & -- & 46.5 & 52.3 & 49.4 & 64.6 & 70.4 & 67.5 & 70.6 & 78.0 & 74.3 & 54.9 & 59.1 & 57.0 \\
GLUS \cite{lin2025glus} & 7B & -- & -- & -- & \underline{48.5} & 54.2 & 51.3 & 65.5 & 69.0 & 67.3 & -- & -- & -- &52.4 & 57.3 & 54.9  \\
Sa2VA \cite{yuan2025sa2va} & 4B & -- & -- & 52.1  & -- & -- & 46.2 & -- & -- & 70.0 & -- & -- & 73.8 & -- & -- & 53.2 \\
RGA3 \cite{wang2025object} & 7B & 56.7 & 62.6 & 59.7  & 47.4 & 52.8 & 50.1 & 66.8 & 70.1 & 68.5  & 68.3 & 77.3 & 72.8 & 55.9 & 60.0 & 58.0 \\
UniPixel \cite{liu2025unipixel} & 7B & \underline{56.9} & \underline{62.9} & \underline{59.9} & \textbf{52.3} & \underline{57.1} & 54.7 & \underline{70.2} & \underline{74.1} & \underline{72.1} & \textbf{71.4} & \textbf{80.0} & \textbf{75.7} & \underline{61.9} & \underline{66.1} & \underline{64.0}  \\
\midrule[0.5pt]
\textbf{\modelgradient} & 7B  & \textbf{60.9} & \textbf{64.6} & \textbf{62.8} & \textbf{52.3} & \textbf{57.8} & \textbf{55.1} & \textbf{72.2} & \textbf{74.6} & \textbf{73.4} & \underline{70.3} & \underline{78.9} & \underline{74.6} & \textbf{62.7} & \textbf{66.9} & \textbf{64.8}  \\
            \bottomrule[1.2pt]
       \end{tabular}
   }
   \vspace{-0.2cm}
   \label{tab:video_seg}
\end{table*}
\begin{table}[htbp]
\setlength{\tabcolsep}{1.2mm} 
\centering
\caption{Experimental results on Charades-STA \cite{gao2017tall} for temporal grounding task, and comparision with state-of-the-art methods. The \textbf{bold} and \underline{underlined} text indicate the best and second-best results.}
\label{tab:charades-sta}
\renewcommand{\arraystretch}{1}
\adjustbox{width=0.97\linewidth}{
   \begin{tabular}{l|c|cccc}
       \toprule[1.2pt]
            \multirow{2}{*}{Model}
            &\multirow{2}{*}{Size}
            &\multicolumn{4}{c}{Charades-STA} \\
    \cmidrule{3-6} 
                & &R@0.3 &R@0.5 &R@0.7 &tIoU   \\
\midrule[0.5pt]  
Video-ChatGPT \cite{maaz2023video} &7B &27.2 &6.2 &1.9 &19.7  \\
VideoChat \cite{li2023videochat} &7B &32.8 &8.6 &0.0 &25.9  \\
Momentor \cite{qian2024momentor} &7B &42.6 &26.6 &11.6 &28.5  \\
GroundingGPT \cite{li2024groundinggpt} &7B &-- &29.6 &11.9 &28.7  \\
Seq2Time \cite{deng2025seq2time} &7B &-- &31.2 &13.7 &--  \\
TimeChat \cite{ren2024timechat} &7B &-- &32.2 &13.4 &--  \\
VTimeLLM \cite{huang2024vtimellm} &13B &55.3 &34.3 &14.7 &34.6  \\
ChatVTG \cite{qu2024chatvtg} &7B &52.7 &33.0 &15.9 &34.9  \\
Grounded-VideoLLM \cite{wang2024grounded} &4B &54.2 &36.4 &19.7 &36.8  \\
InternVideo2.5 \cite{wang2025internvideo2} &7B &-- &43.3 &-- &41.7  \\
LLaVA-ST \cite{li2025llava} &7B & \underline{63.1} & \underline{44.8} & \underline{23.4} & \underline{42.4}  \\
\midrule[0.5pt]
\textbf{\modelgradient} &7B & \textbf{71.2} & \textbf{49.6} & \textbf{24.1} & \textbf{44.7}  \\
            \bottomrule[1.2pt]
       \end{tabular}
   }
   \vspace{-0.4cm}
\end{table}

\begin{table*}[htbp]
\setlength{\tabcolsep}{6pt} 
\centering
\caption{Experimental results on UFVideo-Bench for multi-grained video cooperative understanding tasks. The \textbf{bold} and \underline{underlined} text indicate the best and second-best results.}
\renewcommand{\arraystretch}{1}
\adjustbox{width=1\linewidth}{
   \begin{tabular}{l|c|cccc|ccccc|cccccc}
       \toprule[1.2pt]
            \multirow{2}{*}{Model}
            &\multirow{2}{*}{Size}
            &\multicolumn{4}{c|}{\textbf{PixRQA}}
            &\multicolumn{5}{c|}{\textbf{PixHQA}}
            &\multicolumn{6}{c}{\textbf{PixTRQA}} \\
\cmidrule{3-6} \cmidrule{7-11} \cmidrule{12-17}   
                & & $\mathcal{J}$ & $\mathcal{F}$ & $\mathcal{J\&F}$ & SAvg.
                & $\mathcal{J}$ & $\mathcal{F}$ & $\mathcal{J\&F}$ & SAvg.$_{\text{w T}}$ & SAvg.$_{\text{w/o T}}$
                & tIoU & tIoU@0.5 & $\mathcal{J}$ & $\mathcal{F}$ & $\mathcal{J\&F}$ 
                & SAvg. \\
\midrule[0.5pt]
GPT-4o \cite{hurst2024gpt}  & -- & -- & -- & -- & 2.58 & -- & -- & -- & 4.14 & 4.27 & 17.93 & 10.32 & -- & -- & -- & 3.87  \\
Qwen3-VL \cite{qwen3technicalreport}  & 8B & -- & -- & -- & 2.66 & -- & -- & -- & 4.06 & 4.11 & 41.41 & 38.39 & -- & -- & -- & 3.82  \\
Qwen3-VL \cite{qwen3technicalreport}  & 32B & -- & -- & -- & \underline{2.90} & -- & -- & -- & \textbf{4.28} & \underline{4.34} & \underline{49.34} & \underline{45.16} & -- & -- & -- & \underline{3.94}  \\
\midrule[0.5pt]
\modelgradient & 7B   & \textbf{58.70} & \textbf{47.87} & \textbf{53.39} & \textbf{3.35} & \textbf{46.83} & \textbf{53.00} & \textbf{49.91} & \underline{4.22} & \textbf{4.41} & \textbf{49.64} & \textbf{51.61} & \textbf{31.42} & \textbf{33.08} & \textbf{32.25} & \textbf{4.13}  \\
            \bottomrule[1.2pt]
       \end{tabular}
   }
    \vspace{-0.3cm}
   \label{tab:UFVideo-bench}
\end{table*}

\begin{table}[!htbp]
\setlength{\tabcolsep}{2.5pt} 
\centering 
\caption{Ablation results on task unification and temporal tokens setting for UFVideo. (a) Task unification experiments conducts on MeViS \texttt{val}$^\texttt{u}$, Charades-STA and PixTRQA in UFVideo-Bench. (b) Temporal tokens experiments conducts on Charades-STA datasets.}
\renewcommand{\arraystretch}{1} 
\vspace{-0.1cm}
\begin{minipage}{\linewidth} 
\centering 
\caption*{(a) Task Unification}
\label{tab:ablation_a}
\vspace{-0.2cm}
\tiny 
\resizebox{0.95\linewidth}{!}{
\begin{tabular}{cccc|c|c|ccc}
    \toprule[0.65pt]
        \multirow{2}{*}{Seg.}
        &\multirow{2}{*}{Temp.}
        &\multirow{2}{*}{QA}
        &\multirow{2}{*}{UniB.}
        & \multirow{2}{*}{$\mathcal{J\&F}$ }
        & \multirow{2}{*}{tIoU}
        & \multicolumn{3}{c}{PixTRQA} \\
    \cmidrule{7-9}
    & & & & & & tIoU & $\mathcal{J\&F}$ & SAvg. \\
    \midrule[0.35pt]
    \cmark{} & & & & 52.26 & -- & -- & 29.21 & 2.57 \\
    \cmark{} & & \cmark{} & & 55.20 & -- & -- & 30.23 & 3.83 \\
    & \cmark{} & & & -- & 39.20 & 36.20 & -- & 3.76 \\
    & \cmark{} & \cmark{} & & -- & 41.96 & 38.84 & -- & 3.90 \\
    \midrule[0.35pt]
    \cmark{} & \cmark{} & \cmark{} & \cmark{} & \textbf{62.80} & \textbf{44.72} & \textbf{49.63} & \textbf{32.25} & \textbf{4.13} \\
    \bottomrule[0.65pt]
\end{tabular}
}
\end{minipage}

\vspace{10pt}

\begin{minipage}{\linewidth}
\centering
\caption*{(b) Temporal Tokens}
\vspace{-0.2cm}
\tiny 
\resizebox{0.95\linewidth}{!}{
\begin{tabular}{l|c|c|c|c}
    \toprule[0.5pt]
        \multicolumn{1}{l|}{Token Method}
        &\multicolumn{1}{c|}{Token Size}
        &\multicolumn{1}{c|}{R@0.3}
        &\multicolumn{1}{c|}{tIoU}
        &\multicolumn{1}{c}{$\uparrow$} \\
    \midrule[0.3pt]
    \ding{172} Absolute time number & -- & 44.87  & 30.24 & -  \\
    \ding{173} Relative time token & 50 & 48.31  & 31.55 & - \\
    \ding{174} Relative time token & 100 & 60.27  & 39.19 & \textbf{24.22\%} \\
    \ding{175} Relative time token & 150 & 63.82  & 42.07 & 7.63\% \\
    \bottomrule[0.5pt]
\end{tabular}
}
\end{minipage}
\vspace{-0.3cm}
\end{table}

\subsection{Training Strategy}

To effectively integrate different video understanding tasks, we implement a two-phase training approach. In stage 1, we align temporal grounding and segmentation tasks using 58K temporal grounding data and 70K segmentation data to enable the model to align temporal information and train the mask decoder to learn the LLM output embedding into the mask. During this stage, a peak learning rate of 2e-5 is applied to train the visual projector, LLM and mask decoder. 
In stage 2, we utilize approximately 3 million data, effectively integrating the model's understanding of each task and enhancing its performance. We employ the same learning rate as in stage 1 to train the visual projector, region encoder, LLM, and mask decoder in stage 2.

\subsection{Main Results}
\label{sec:main_results}

To evaluate the efficacy of UFVideo, we conduct comprehensive experiments across various common understanding tasks and video cooperative understanding tasks.

\subsubsection{Global Video Understanding}

We evaluate UFVideo's general understanding ability of videos on MVBench \cite{li2024mvbench}. As shown in~\cref{tab:mvbench}, UFVideo achieves top performance in most metrics, its average metric exceeding GPT-4v's by 54.7\% and outperforms Qwen2-VL-7B's performance. These results demonstrate that our model effectively integrates multiple fine-grained video understanding tasks while maintaining and even enhancing its holistic video understanding capability.

\subsubsection{Pixel-Level Video Understanding}

To evaluate pixel-level video understanding, we compare the UFVideo with state-of-the-art methods on video object referring tasks and reasoning and referring video object segmentation.

\vspace{-0.35cm}
\paragraph{Video Object Referring}
We evaluate the object region understanding ability of UFVideo on VideoRefer-Bench, which includes description and question answering tasks. As shown in~\cref{tab:videorefer-d} and ~\cref{tab:videorefer-q}, our model significantly outperforms VideoRefer and UniPixel on both VideoRefer-Bench-D and VideoRefer-Bench-Q \cite{yuan2025videorefer}. Although our model is based on VideoRefer, our results gain 8.04\% improvement on VideoRefer-Bench-Q and 4.34\% in VideoRefer-Bench-Q both in multi-frame mode. These results demonstrate that learning more comprehensive multi-grained tasks enhances UFVideo's understanding in object referring tasks.
\vspace{-0.3cm}
\paragraph{Referring and Reasoning Video Segmentation.}

We utilize the MeViS \cite{ding2023mevis} (\texttt{val}), MeViS \cite{ding2023mevis} (\texttt{val$^\texttt{u}$}), Ref-YouTube-VOS \cite{seo2020urvos} (\texttt{val}) and Ref-DAVIS17 \cite{pont20172017} (\texttt{val}) datasets to assess UFVideo's performance in the referring video object segmentation task. Additionally, we use the ReVOS \cite{yan2024visa} \texttt{val} dataset to assess UFVideo's capability in reasoning video object segmentation, which involves implicit instructions requiring a deeper video understanding compared to referring segmentation. As shown in~\cref{tab:video_seg}, UFVideo delivers impressive results, achieving top performance across nearly all benchmarks, especially on the challenging ReVOS benchmarks, where UFVideo surpasses RGA3 by 11.7\%. This highlights the synergistic effect of integrating multiple granularity video information.

\subsubsection{Temporal Video Grounding.}

As shown in~\cref{tab:charades-sta}, we evaluate our capability for temporal event localization within videos using the Charades-STA benchmark \cite{gao2017tall}. UFVideo is the only model in the table that unifies fine-grained pixel-level and temporal-level understanding, outperforming previous Video LLMs specialized in temporal video grounding, LLaVA-ST, achieving an improvement of 5.42\% in tIoU. This demonstrates the effectiveness of our designed temporal token.

\subsubsection{Multi-grained Video Understanding}

We will evaluate the effectiveness of UFVideo on multi-grained video cooperative understanding task. As our model is currently the only Video LLM supporting comprehensive cooperative tasks, we have chosen two strong large multi-modal models GPT-4o \cite{hurst2024gpt}, Qwen3-VL-8B-Instruct and Qwen3-VL-32B-Instruct \cite{qwen3technicalreport} as baselines. 
As shown in~\cref{tab:UFVideo-bench}, UFVideo achieves state-of-the-art performance on cooperative understanding tasks. It outperforms the similarly sized Qwen3-VL-8B in semantic accuracy and temporal grounding, and shows a notable advantage comparable to Qwen3-VL-32B. UFVideo demonstrates the capability to flexibly handle multi-grained video cooperative understanding tasks, and the benefits of joint multi-grained tasks learning, while also highlighting the complexity of video cooperative understanding tasks.

\subsection{Ablation Study}
\label{ablation_study}
\paragraph{Ablation on Task Unification.}

~\cref{tab:ablation_a} (a) illustrates the model's unified effectiveness across various fine-grained video understanding tasks. We explore the model's performance when segmentation and temporal grounding tasks are added, and further enhance the capabilities of other tasks by introducing QA data for joint training when new segmentation or temporal grounding tasks are added. In the final row, where all tasks are trained together, this demonstrates that the model not only achieves flexibly unification of multiple tasks but also promotes mutual enhancement of understanding among them.

\vspace{-0.4cm}

\paragraph{Ablation on Temporal Tokens.}

To effectively incorporate temporal grounding capabilities, we compared the performance of LLMs using absolute temporal numbers, relative temporal tokens, and varying token counts under the same experimental setup. As shown in~\cref{tab:ablation_a} (b), relative temporal tokens outperform absolute temporal formats. Moreover, while more tokens generally improve performance, the improvement between 150 and 100 tokens is less significant than the improvement between 100 and 50 tokens. Thus, we opted for 100 tokens to balance computational cost and effectiveness.

\vspace{-0.1cm}
\vspace{-0.1cm}
\section{Conclusion}

In this work, we proposed UFVideo, a Video LLM that achieves unified fine-grained video understanding tasks, including global, pixel-level referring and segmentation, and temporal-level grounding tasks. We design a unified visual-language guided alignment to efficiently handle all tasks. Experimental results prove that joint multi-grained information can mutually enhance understanding. We also introduced UFVideo-Bench to address the limitations of multi-grained video cooperative understanding tasks. Extensive experiments on UFVideo-Bench reveal the complexity of video cooperative understanding and the significance of our model. We hope this work can provides valuable insights for advancing future Video LLMs. 

\section{Acknowledgment}

This work is supported by the National Natural Science Foundation of China (Grant No.62572206). Minghui Li is the corresponding author.
{
    \small
    \bibliographystyle{ieeenat_fullname}
    \bibliography{main}

@String(ICASSP=	{ICASSP})

@String(AAAI = {AAAI})

@article{liu2023visual,
  title={Visual instruction tuning},
  author={Liu, Haotian and Li, Chunyuan and Wu, Qingyang and Lee, Yong Jae},
  journal={Advances in neural information processing systems},
  volume={36},
  pages={34892--34916},
  year={2023}
}

@article{huang2025step,
  title={Step-audio: Unified understanding and generation in intelligent speech interaction},
  author={Huang, Ailin and Wu, Boyong and Wang, Bruce and Yan, Chao and Hu, Chen and Feng, Chengli and Tian, Fei and Shen, Feiyu and Li, Jingbei and Chen, Mingrui and others},
  journal={arXiv preprint arXiv:2502.11946},
  year={2025}
}

@article{hurst2024gpt,
  title={Gpt-4o system card},
  author={Hurst, Aaron and Lerer, Adam and Goucher, Adam P and Perelman, Adam and Ramesh, Aditya and Clark, Aidan and Ostrow, AJ and Welihinda, Akila and Hayes, Alan and Radford, Alec and others},
  journal={arXiv preprint arXiv:2410.21276},
  year={2024}
}

@article{maaz2023video,
  title={Video-chatgpt: Towards detailed video understanding via large vision and language models},
  author={Maaz, Muhammad and Rasheed, Hanoona and Khan, Salman and Khan, Fahad Shahbaz},
  journal={arXiv preprint arXiv:2306.05424},
  year={2023}
}

@article{bai2025qwen2,
  title={Qwen2. 5-vl technical report},
  author={Bai, Shuai and Chen, Keqin and Liu, Xuejing and Wang, Jialin and Ge, Wenbin and Song, Sibo and Dang, Kai and Wang, Peng and Wang, Shijie and Tang, Jun and others},
  journal={arXiv preprint arXiv:2502.13923},
  year={2025}
}

@inproceedings{yuan2025videorefer,
  title={Videorefer suite: Advancing spatial-temporal object understanding with video llm},
  author={Yuan, Yuqian and Zhang, Hang and Li, Wentong and Cheng, Zesen and Zhang, Boqiang and Li, Long and Li, Xin and Zhao, Deli and Zhang, Wenqiao and Zhuang, Yueting and others},
  booktitle={Proceedings of the Computer Vision and Pattern Recognition Conference},
  pages={18970--18980},
  year={2025}
}

@inproceedings{li2025llava,
  title={Llava-st: A multimodal large language model for fine-grained spatial-temporal understanding},
  author={Li, Hongyu and Chen, Jinyu and Wei, Ziyu and Huang, Shaofei and Hui, Tianrui and Gao, Jialin and Wei, Xiaoming and Liu, Si},
  booktitle={Proceedings of the Computer Vision and Pattern Recognition Conference},
  pages={8592--8603},
  year={2025}
}

@article{lin2025perceive,
  title={Perceive Anything: Recognize, Explain, Caption, and Segment Anything in Images and Videos},
  author={Lin, Weifeng and Wei, Xinyu and An, Ruichuan and Ren, Tianhe and Chen, Tingwei and Zhang, Renrui and Guo, Ziyu and Zhang, Wentao and Zhang, Lei and Li, Hongsheng},
  journal={arXiv preprint arXiv:2506.05302},
  year={2025}
}

@article{wei2024instructseg,
  title={Instructseg: Unifying instructed visual segmentation with multi-modal large language models},
  author={Wei, Cong and Zhong, Yujie and Tan, Haoxian and Zeng, Yingsen and Liu, Yong and Zhao, Zheng and Yang, Yujiu},
  journal={arXiv preprint arXiv:2412.14006},
  year={2024}
}

@article{zeng2025distime,
  title={DisTime: Distribution-based Time Representation for Video Large Language Models},
  author={Zeng, Yingsen and Huang, Zepeng and Zhong, Yujie and Feng, Chengjian and Hu, Jie and Ma, Lin and Liu, Yang},
  journal={arXiv preprint arXiv:2505.24329},
  year={2025}
}

@article{zhang2025videollama,
  title={Videollama 3: Frontier multimodal foundation models for image and video understanding},
  author={Zhang, Boqiang and Li, Kehan and Cheng, Zesen and Hu, Zhiqiang and Yuan, Yuqian and Chen, Guanzheng and Leng, Sicong and Jiang, Yuming and Zhang, Hang and Li, Xin and others},
  journal={arXiv preprint arXiv:2501.13106},
  year={2025}
}

@inproceedings{zhang2024gpt4roi,
  title={Gpt4roi: Instruction tuning large language model on region-of-interest},
  author={Zhang, Shilong and Sun, Peize and Chen, Shoufa and Xiao, Min and Shao, Wenqi and Zhang, Wenwei and Liu, Yu and Chen, Kai and Luo, Ping},
  booktitle={European conference on computer vision},
  pages={52--70},
  year={2024},
  organization={Springer}
}

@article{wang2024cogvlm,
  title={Cogvlm: Visual expert for pretrained language models},
  author={Wang, Weihan and Lv, Qingsong and Yu, Wenmeng and Hong, Wenyi and Qi, Ji and Wang, Yan and Ji, Junhui and Yang, Zhuoyi and Zhao, Lei and XiXuan, Song and others},
  journal={Advances in Neural Information Processing Systems},
  volume={37},
  pages={121475--121499},
  year={2024}
}

@article{peng2023kosmos,
  title={Kosmos-2: Grounding multimodal large language models to the world},
  author={Peng, Zhiliang and Wang, Wenhui and Dong, Li and Hao, Yaru and Huang, Shaohan and Ma, Shuming and Wei, Furu},
  journal={arXiv preprint arXiv:2306.14824},
  year={2023}
}

@article{chen2023shikra,
  title={Shikra: Unleashing multimodal llm's referential dialogue magic},
  author={Chen, Keqin and Zhang, Zhao and Zeng, Weili and Zhang, Richong and Zhu, Feng and Zhao, Rui},
  journal={arXiv preprint arXiv:2306.15195},
  year={2023}
}

@article{lian2025describe,
  title={Describe anything: Detailed localized image and video captioning},
  author={Lian, Long and Ding, Yifan and Ge, Yunhao and Liu, Sifei and Mao, Hanzi and Li, Boyi and Pavone, Marco and Liu, Ming-Yu and Darrell, Trevor and Yala, Adam and others},
  journal={arXiv preprint arXiv:2504.16072},
  year={2025}
}

@article{sun2025sama,
  title={SAMA: Towards Multi-Turn Referential Grounded Video Chat with Large Language Models},
  author={Sun, Ye and Zhang, Hao and Ding, Henghui and Zhang, Tiehua and Ma, Xingjun and Jiang, Yu-Gang},
  journal={arXiv preprint arXiv:2505.18812},
  year={2025}
}

@inproceedings{huang2024vtimellm,
  title={Vtimellm: Empower llm to grasp video moments},
  author={Huang, Bin and Wang, Xin and Chen, Hong and Song, Zihan and Zhu, Wenwu},
  booktitle={Proceedings of the IEEE/CVF Conference on Computer Vision and Pattern Recognition},
  pages={14271--14280},
  year={2024}
}

@article{huang2020multimodal,
  title={Multimodal pretraining for dense video captioning},
  author={Huang, Gabriel and Pang, Bo and Zhu, Zhenhai and Rivera, Clara and Soricut, Radu},
  journal={arXiv preprint arXiv:2011.11760},
  year={2020}
}

@inproceedings{tang2019coin,
  title={Coin: A large-scale dataset for comprehensive instructional video analysis},
  author={Tang, Yansong and Ding, Dajun and Rao, Yongming and Zheng, Yu and Zhang, Danyang and Zhao, Lili and Lu, Jiwen and Zhou, Jie},
  booktitle={Proceedings of the IEEE/CVF Conference on Computer Vision and Pattern Recognition},
  pages={1207--1216},
  year={2019}
}

@article{wang2024grounded,
  title={Grounded-videollm: Sharpening fine-grained temporal grounding in video large language models},
  author={Wang, Haibo and Xu, Zhiyang and Cheng, Yu and Diao, Shizhe and Zhou, Yufan and Cao, Yixin and Wang, Qifan and Ge, Weifeng and Huang, Lifu},
  journal={arXiv preprint arXiv:2410.03290},
  year={2024}
}

@inproceedings{lai2024lisa,
  title={Lisa: Reasoning segmentation via large language model},
  author={Lai, Xin and Tian, Zhuotao and Chen, Yukang and Li, Yanwei and Yuan, Yuhui and Liu, Shu and Jia, Jiaya},
  booktitle={Proceedings of the IEEE/CVF Conference on Computer Vision and Pattern Recognition},
  pages={9579--9589},
  year={2024}
}

@inproceedings{pi2024perceptiongpt,
  title={Perceptiongpt: Effectively fusing visual perception into llm},
  author={Pi, Renjie and Yao, Lewei and Gao, Jiahui and Zhang, Jipeng and Zhang, Tong},
  booktitle={Proceedings of the IEEE/CVF conference on computer vision and pattern recognition},
  pages={27124--27133},
  year={2024}
}

@inproceedings{ren2024pixellm,
  title={Pixellm: Pixel reasoning with large multimodal model},
  author={Ren, Zhongwei and Huang, Zhicheng and Wei, Yunchao and Zhao, Yao and Fu, Dongmei and Feng, Jiashi and Jin, Xiaojie},
  booktitle={Proceedings of the IEEE/CVF Conference on Computer Vision and Pattern Recognition},
  pages={26374--26383},
  year={2024}
}

@inproceedings{xia2024gsva,
  title={Gsva: Generalized segmentation via multimodal large language models},
  author={Xia, Zhuofan and Han, Dongchen and Han, Yizeng and Pan, Xuran and Song, Shiji and Huang, Gao},
  booktitle={Proceedings of the IEEE/CVF Conference on Computer Vision and Pattern Recognition},
  pages={3858--3869},
  year={2024}
}

@inproceedings{yan2024visa,
  title={Visa: Reasoning video object segmentation via large language models},
  author={Yan, Cilin and Wang, Haochen and Yan, Shilin and Jiang, Xiaolong and Hu, Yao and Kang, Guoliang and Xie, Weidi and Gavves, Efstratios},
  booktitle={European Conference on Computer Vision},
  pages={98--115},
  year={2024},
  organization={Springer}
}

@article{wang2024qwen2,
  title={Qwen2-vl: Enhancing vision-language model's perception of the world at any resolution},
  author={Wang, Peng and Bai, Shuai and Tan, Sinan and Wang, Shijie and Fan, Zhihao and Bai, Jinze and Chen, Keqin and Liu, Xuejing and Wang, Jialin and Ge, Wenbin and others},
  journal={arXiv preprint arXiv:2409.12191},
  year={2024}
}

@article{li2023videochat,
  title={Videochat: Chat-centric video understanding},
  author={Li, KunChang and He, Yinan and Wang, Yi and Li, Yizhuo and Wang, Wenhai and Luo, Ping and Wang, Yali and Wang, Limin and Qiao, Yu},
  journal={arXiv preprint arXiv:2305.06355},
  year={2023}
}

@article{qiu2024artemis,
  title={Artemis: Towards referential understanding in complex videos},
  author={Qiu, Jihao and Zhang, Yuan and Tang, Xi and Xie, Lingxi and Ma, Tianren and Yan, Pengyu and Doermann, David and Ye, Qixiang and Tian, Yunjie},
  journal={Advances in Neural Information Processing Systems},
  volume={37},
  pages={114321--114347},
  year={2024}
}

@article{li2024groundinggpt,
  title={Groundinggpt: Language enhanced multi-modal grounding model},
  author={Li, Zhaowei and Xu, Qi and Zhang, Dong and Song, Hang and Cai, Yiqing and Qi, Qi and Zhou, Ran and Pan, Junting and Li, Zefeng and Vu, Van Tu and others},
  journal={arXiv preprint arXiv:2401.06071},
  year={2024}
}

@article{wei2024hyperseg,
  title={Hyperseg: Towards universal visual segmentation with large language model},
  author={Wei, Cong and Zhong, Yujie and Tan, Haoxian and Liu, Yong and Zhao, Zheng and Hu, Jie and Yang, Yujiu},
  journal={arXiv preprint arXiv:2411.17606},
  year={2024}
}

@article{cheng2024videollama,
  title={Videollama 2: Advancing spatial-temporal modeling and audio understanding in video-llms},
  author={Cheng, Zesen and Leng, Sicong and Zhang, Hang and Xin, Yifei and Li, Xin and Chen, Guanzheng and Zhu, Yongxin and Zhang, Wenqi and Luo, Ziyang and Zhao, Deli and others},
  journal={arXiv preprint arXiv:2406.07476},
  year={2024}
}

@inproceedings{li2024mvbench,
  title={Mvbench: A comprehensive multi-modal video understanding benchmark},
  author={Li, Kunchang and Wang, Yali and He, Yinan and Li, Yizhuo and Wang, Yi and Liu, Yi and Wang, Zun and Xu, Jilan and Chen, Guo and Luo, Ping and others},
  booktitle={Proceedings of the IEEE/CVF Conference on Computer Vision and Pattern Recognition},
  pages={22195--22206},
  year={2024}
}

@article{chen2024sharegpt4video,
  title={Sharegpt4video: Improving video understanding and generation with better captions},
  author={Chen, Lin and Wei, Xilin and Li, Jinsong and Dong, Xiaoyi and Zhang, Pan and Zang, Yuhang and Chen, Zehui and Duan, Haodong and Tang, Zhenyu and Yuan, Li and others},
  journal={Advances in Neural Information Processing Systems},
  volume={37},
  pages={19472--19495},
  year={2024}
}

@article{liu2025unipixel,
  title={UniPixel: Unified Object Referring and Segmentation for Pixel-Level Visual Reasoning},
  author={Liu, Ye and Ma, Zongyang and Pu, Junfu and Qi, Zhongang and Wu, Yang and Shan, Ying and Chen, Chang Wen},
  journal={arXiv preprint arXiv:2509.18094},
  year={2025}
}

@article{wang2025object,
  title={Object-centric video question answering with visual grounding and referring},
  author={Wang, Haochen and Chen, Qirui and Yan, Cilin and Cai, Jiayin and Jiang, Xiaolong and Hu, Yao and Xie, Weidi and Gavves, Stratis},
  journal={arXiv preprint arXiv:2507.19599},
  year={2025}
}

@article{you2023ferret,
  title={Ferret: Refer and ground anything anywhere at any granularity},
  author={You, Haoxuan and Zhang, Haotian and Gan, Zhe and Du, Xianzhi and Zhang, Bowen and Wang, Zirui and Cao, Liangliang and Chang, Shih-Fu and Yang, Yinfei},
  journal={arXiv preprint arXiv:2310.07704},
  year={2023}
}

@inproceedings{gao2017tall,
  title={Tall: Temporal activity localization via language query},
  author={Gao, Jiyang and Sun, Chen and Yang, Zhenheng and Nevatia, Ram},
  booktitle={Proceedings of the IEEE international conference on computer vision},
  pages={5267--5275},
  year={2017}
}

@inproceedings{grauman2022ego4d,
  title={Ego4d: Around the world in 3,000 hours of egocentric video},
  author={Grauman, Kristen and Westbury, Andrew and Byrne, Eugene and Chavis, Zachary and Furnari, Antonino and Girdhar, Rohit and Hamburger, Jackson and Jiang, Hao and Liu, Miao and Liu, Xingyu and others},
  booktitle={Proceedings of the IEEE/CVF conference on computer vision and pattern recognition},
  pages={18995--19012},
  year={2022}
}

@inproceedings{xiao2024can,
  title={Can i trust your answer? visually grounded video question answering},
  author={Xiao, Junbin and Yao, Angela and Li, Yicong and Chua, Tat-Seng},
  booktitle={Proceedings of the IEEE/CVF Conference on Computer Vision and Pattern Recognition},
  pages={13204--13214},
  year={2024}
}

@inproceedings{heo2025omni,
  title={Omni-rgpt: Unifying image and video region-level understanding via token marks},
  author={Heo, Miran and Chen, Min-Hung and Huang, De-An and Liu, Sifei and Radhakrishnan, Subhashree and Kim, Seon Joo and Wang, Yu-Chiang Frank and Hachiuma, Ryo},
  booktitle={Proceedings of the Computer Vision and Pattern Recognition Conference},
  pages={3919--3930},
  year={2025}
}

@misc{qwen3technicalreport,
      title={Qwen3 Technical Report}, 
      author={Qwen Team},
      year={2025},
      eprint={2505.09388},
      archivePrefix={arXiv},
      primaryClass={cs.CL},
      url={https://arxiv.org/abs/2505.09388}, 
}

@misc{zhai2023sigmoid,
      title={Sigmoid Loss for Language Image Pre-Training}, 
      author={Xiaohua Zhai and Basil Mustafa and Alexander Kolesnikov and Lucas Beyer},
      year={2023},
      eprint={2303.15343},
      archivePrefix={arXiv},
      primaryClass={cs.CV}
}

@article{ravi2024sam,
  title={Sam 2: Segment anything in images and videos},
  author={Ravi, Nikhila and Gabeur, Valentin and Hu, Yuan-Ting and Hu, Ronghang and Ryali, Chaitanya and Ma, Tengyu and Khedr, Haitham and R{\"a}dle, Roman and Rolland, Chloe and Gustafson, Laura and others},
  journal={arXiv preprint arXiv:2408.00714},
  year={2024}
}

@inproceedings{fu2025video,
  title={Video-mme: The first-ever comprehensive evaluation benchmark of multi-modal llms in video analysis},
  author={Fu, Chaoyou and Dai, Yuhan and Luo, Yongdong and Li, Lei and Ren, Shuhuai and Zhang, Renrui and Wang, Zihan and Zhou, Chenyu and Shen, Yunhang and Zhang, Mengdan and others},
  booktitle={Proceedings of the Computer Vision and Pattern Recognition Conference},
  pages={24108--24118},
  year={2025}
}

@inproceedings{krishna2017dense,
  title={Dense-captioning events in videos},
  author={Krishna, Ranjay and Hata, Kenji and Ren, Frederic and Fei-Fei, Li and Carlos Niebles, Juan},
  booktitle={Proceedings of the IEEE international conference on computer vision},
  pages={706--715},
  year={2017}
}

@article{li2024llava,
  title={Llava-onevision: Easy visual task transfer},
  author={Li, Bo and Zhang, Yuanhan and Guo, Dong and Zhang, Renrui and Li, Feng and Zhang, Hao and Zhang, Kaichen and Zhang, Peiyuan and Li, Yanwei and Liu, Ziwei and others},
  journal={arXiv preprint arXiv:2408.03326},
  year={2024}
}

@misc{team2024internvl2,
  title={Internvl2: Better than the best—expanding performance boundaries of open-source multimodal models with the progressive scaling strategy},
  author={Team, OpenGVLab},
  year={2024},
  publisher={Accessed}
}

@inproceedings{yuan2024osprey,
  title={Osprey: Pixel understanding with visual instruction tuning},
  author={Yuan, Yuqian and Li, Wentong and Liu, Jian and Tang, Dongqi and Luo, Xinjie and Qin, Chi and Zhang, Lei and Zhu, Jianke},
  booktitle={Proceedings of the IEEE/CVF Conference on Computer Vision and Pattern Recognition},
  pages={28202--28211},
  year={2024}
}

@article{qian2024momentor,
  title={Momentor: Advancing video large language model with fine-grained temporal reasoning},
  author={Qian, Long and Li, Juncheng and Wu, Yu and Ye, Yaobo and Fei, Hao and Chua, Tat-Seng and Zhuang, Yueting and Tang, Siliang},
  journal={arXiv preprint arXiv:2402.11435},
  year={2024}
}

@inproceedings{deng2025seq2time,
  title={Seq2time: Sequential knowledge transfer for video llm temporal grounding},
  author={Deng, Andong and Gao, Zhongpai and Choudhuri, Anwesa and Planche, Benjamin and Zheng, Meng and Wang, Bin and Chen, Terrence and Chen, Chen and Wu, Ziyan},
  booktitle={Proceedings of the Computer Vision and Pattern Recognition Conference},
  pages={13766--13775},
  year={2025}
}

@inproceedings{ren2024timechat,
  title={Timechat: A time-sensitive multimodal large language model for long video understanding},
  author={Ren, Shuhuai and Yao, Linli and Li, Shicheng and Sun, Xu and Hou, Lu},
  booktitle={Proceedings of the IEEE/CVF Conference on Computer Vision and Pattern Recognition},
  pages={14313--14323},
  year={2024}
}

@inproceedings{qu2024chatvtg,
  title={Chatvtg: Video temporal grounding via chat with video dialogue large language models},
  author={Qu, Mengxue and Chen, Xiaodong and Liu, Wu and Li, Alicia and Zhao, Yao},
  booktitle={Proceedings of the IEEE/CVF Conference on Computer Vision and Pattern Recognition},
  pages={1847--1856},
  year={2024}
}

@article{wang2025internvideo2,
  title={Internvideo2. 5: Empowering video mllms with long and rich context modeling},
  author={Wang, Yi and Li, Xinhao and Yan, Ziang and He, Yinan and Yu, Jiashuo and Zeng, Xiangyu and Wang, Chenting and Ma, Changlian and Huang, Haian and Gao, Jianfei and others},
  journal={arXiv preprint arXiv:2501.12386},
  year={2025}
}

@inproceedings{wu2022language,
  title={Language as queries for referring video object segmentation},
  author={Wu, Jiannan and Jiang, Yi and Sun, Peize and Yuan, Zehuan and Luo, Ping},
  booktitle={Proceedings of the IEEE/CVF Conference on Computer Vision and Pattern Recognition},
  pages={4974--4984},
  year={2022}
}

@inproceedings{ding2023mevis,
  title={MeViS: A large-scale benchmark for video segmentation with motion expressions},
  author={Ding, Henghui and Liu, Chang and He, Shuting and Jiang, Xudong and Loy, Chen Change},
  booktitle={Proceedings of the IEEE/CVF international conference on computer vision},
  pages={2694--2703},
  year={2023}
}

@inproceedings{wu2023onlinerefer,
  title={Onlinerefer: A simple online baseline for referring video object segmentation},
  author={Wu, Dongming and Wang, Tiancai and Zhang, Yuang and Zhang, Xiangyu and Shen, Jianbing},
  booktitle={Proceedings of the IEEE/CVF International Conference on Computer Vision},
  pages={2761--2770},
  year={2023}
}

@article{bai2024one,
  title={One token to seg them all: Language instructed reasoning segmentation in videos},
  author={Bai, Zechen and He, Tong and Mei, Haiyang and Wang, Pichao and Gao, Ziteng and Chen, Joya and Zhang, Zheng and Shou, Mike Zheng},
  journal={Advances in Neural Information Processing Systems},
  volume={37},
  pages={6833--6859},
  year={2024}
}

@inproceedings{munasinghe2025videoglamm,
  title={Videoglamm: A large multimodal model for pixel-level visual grounding in videos},
  author={Munasinghe, Shehan and Gani, Hanan and Zhu, Wenqi and Cao, Jiale and Xing, Eric and Khan, Fahad Shahbaz and Khan, Salman},
  booktitle={Proceedings of the Computer Vision and Pattern Recognition Conference},
  pages={19036--19046},
  year={2025}
}

@inproceedings{zheng2025villa,
  title={Villa: Video reasoning segmentation with large language model},
  author={Zheng, Rongkun and Qi, Lu and Chen, Xi and Wang, Yi and Wang, Kun and Zhao, Hengshuang},
  booktitle={Proceedings of the IEEE/CVF International Conference on Computer Vision},
  pages={23667--23677},
  year={2025}
}

@inproceedings{lin2025glus,
  title={Glus: Global-local reasoning unified into a single large language model for video segmentation},
  author={Lin, Lang and Yu, Xueyang and Pang, Ziqi and Wang, Yu-Xiong},
  booktitle={Proceedings of the Computer Vision and Pattern Recognition Conference},
  pages={8658--8667},
  year={2025}
}

@article{yuan2025sa2va,
  title={Sa2va: Marrying sam2 with llava for dense grounded understanding of images and videos},
  author={Yuan, Haobo and Li, Xiangtai and Zhang, Tao and Huang, Zilong and Xu, Shilin and Ji, Shunping and Tong, Yunhai and Qi, Lu and Feng, Jiashi and Yang, Ming-Hsuan},
  journal={arXiv preprint arXiv:2501.04001},
  year={2025}
}

@inproceedings{botach2022end,
  title={End-to-end referring video object segmentation with multimodal transformers},
  author={Botach, Adam and Zheltonozhskii, Evgenii and Baskin, Chaim},
  booktitle={Proceedings of the IEEE/CVF Conference on Computer Vision and Pattern Recognition},
  pages={4985--4995},
  year={2022}
}

@article{stroh2024trackgpt,
  title={TrackGPT--A generative pre-trained transformer for cross-domain entity trajectory forecasting},
  author={Stroh, Nicholas},
  journal={arXiv preprint arXiv:2402.00066},
  year={2024}
}

@inproceedings{caba2015activitynet,
  title={Activitynet: A large-scale video benchmark for human activity understanding},
  author={Caba Heilbron, Fabian and Escorcia, Victor and Ghanem, Bernard and Carlos Niebles, Juan},
  booktitle={Proceedings of the ieee conference on computer vision and pattern recognition},
  pages={961--970},
  year={2015}
}

@inproceedings{zhou2018towards,
  title={Towards automatic learning of procedures from web instructional videos},
  author={Zhou, Luowei and Xu, Chenliang and Corso, Jason},
  booktitle={Proceedings of the AAAI conference on artificial intelligence},
  volume={32},
  number={1},
  year={2018}
}

@article{patraucean2023perception,
  title={Perception test: A diagnostic benchmark for multimodal video models},
  author={Patraucean, Viorica and Smaira, Lucas and Gupta, Ankush and Recasens, Adria and Markeeva, Larisa and Banarse, Dylan and Koppula, Skanda and Malinowski, Mateusz and Yang, Yi and Doersch, Carl and others},
  journal={Advances in Neural Information Processing Systems},
  volume={36},
  pages={42748--42761},
  year={2023}
}

@article{xu2024pllava,
  title={Pllava: Parameter-free llava extension from images to videos for video dense captioning},
  author={Xu, Lin and Zhao, Yilin and Zhou, Daquan and Lin, Zhijie and Ng, See Kiong and Feng, Jiashi},
  journal={arXiv preprint arXiv:2404.16994},
  year={2024}
}

@article{maaz2024videogpt+,
  title={Videogpt+: Integrating image and video encoders for enhanced video understanding},
  author={Maaz, Muhammad and Rasheed, Hanoona and Khan, Salman and Khan, Fahad},
  journal={arXiv preprint arXiv:2406.09418},
  year={2024}
}

@misc{openai2023gpt4v,
  author    = {OpenAI},
  title     = {GPT-4V},
  howpublished = {\url{https://openai.com/index/gpt-4v-system-card/}},
  year      = {2023}
}

@inproceedings{lin2024video,
  title={Video-llava: Learning united visual representation by alignment before projection},
  author={Lin, Bin and Ye, Yang and Zhu, Bin and Cui, Jiaxi and Ning, Munan and Jin, Peng and Yuan, Li},
  booktitle={Proceedings of the 2024 Conference on Empirical Methods in Natural Language Processing},
  pages={5971--5984},
  year={2024}
}

@inproceedings{seo2020urvos,
  title={Urvos: Unified referring video object segmentation network with a large-scale benchmark},
  author={Seo, Seonguk and Lee, Joon-Young and Han, Bohyung},
  booktitle={European conference on computer vision},
  pages={208--223},
  year={2020},
  organization={Springer}
}

@article{pont20172017,
  title={The 2017 davis challenge on video object segmentation},
  author={Pont-Tuset, Jordi and Perazzi, Federico and Caelles, Sergi and Arbel{\'a}ez, Pablo and Sorkine-Hornung, Alex and Van Gool, Luc},
  journal={arXiv preprint arXiv:1704.00675},
  year={2017}
}

@inproceedings{li2023blip,
  title={Blip-2: Bootstrapping language-image pre-training with frozen image encoders and large language models},
  author={Li, Junnan and Li, Dongxu and Savarese, Silvio and Hoi, Steven},
  booktitle={International conference on machine learning},
  pages={19730--19742},
  year={2023},
  organization={PMLR}
}

@article{wang2403internvideo2,
  title={Internvideo2: Scaling video foundation models for multimodal video understanding. arXiv 2024},
  author={Wang, Y and Li, K and Li, X and Yu, J and He, Y and Chen, G and Pei, B and Zheng, R and Xu, J and Wang, Z and others},
  journal={arXiv preprint arXiv:2403.15377}
}

@inproceedings{huang2024lita,
  title={Lita: Language instructed temporal-localization assistant},
  author={Huang, De-An and Liao, Shijia and Radhakrishnan, Subhashree and Yin, Hongxu and Molchanov, Pavlo and Yu, Zhiding and Kautz, Jan},
  booktitle={European Conference on Computer Vision},
  pages={202--218},
  year={2024},
  organization={Springer}
}

@inproceedings{anne2017localizing,
  title={Localizing moments in video with natural language},
  author={Anne Hendricks, Lisa and Wang, Oliver and Shechtman, Eli and Sivic, Josef and Darrell, Trevor and Russell, Bryan},
  booktitle={Proceedings of the IEEE international conference on computer vision},
  pages={5803--5812},
  year={2017}
}

@inproceedings{zala2023hierarchical,
  title={Hierarchical video-moment retrieval and step-captioning},
  author={Zala, Abhay and Cho, Jaemin and Kottur, Satwik and Chen, Xilun and Oguz, Barlas and Mehdad, Yashar and Bansal, Mohit},
  booktitle={Proceedings of the IEEE/CVF Conference on Computer Vision and Pattern Recognition},
  pages={23056--23065},
  year={2023}
}

@inproceedings{oncescu2021queryd,
  title={Queryd: A video dataset with high-quality text and audio narrations},
  author={Oncescu, Andreea-Maria and Henriques, Joao F and Liu, Yang and Zisserman, Andrew and Albanie, Samuel},
  booktitle={ICASSP 2021-2021 IEEE International Conference on Acoustics, Speech and Signal Processing (ICASSP)},
  pages={2265--2269},
  year={2021},
  organization={IEEE}
}

@inproceedings{guo2025vtg,
  title={Vtg-llm: Integrating timestamp knowledge into video llms for enhanced video temporal grounding},
  author={Guo, Yongxin and Liu, Jingyu and Li, Mingda and Cheng, Dingxin and Tang, Xiaoying and Sui, Dianbo and Liu, Qingbin and Chen, Xi and Zhao, Kevin},
  booktitle={Proceedings of the AAAI Conference on Artificial Intelligence},
  volume={39},
  number={3},
  pages={3302--3310},
  year={2025}
}

@inproceedings{xiao2021next,
  title={Next-qa: Next phase of question-answering to explaining temporal actions},
  author={Xiao, Junbin and Shang, Xindi and Yao, Angela and Chua, Tat-Seng},
  booktitle={Proceedings of the IEEE/CVF conference on computer vision and pattern recognition},
  pages={9777--9786},
  year={2021}
}

@inproceedings{li2023intentqa,
  title={Intentqa: Context-aware video intent reasoning},
  author={Li, Jiapeng and Wei, Ping and Han, Wenjuan and Fan, Lifeng},
  booktitle={Proceedings of the IEEE/CVF international conference on computer vision},
  pages={11963--11974},
  year={2023}
}

@inproceedings{grunde2021agqa,
  title={Agqa: A benchmark for compositional spatio-temporal reasoning},
  author={Grunde-McLaughlin, Madeleine and Krishna, Ranjay and Agrawala, Maneesh},
  booktitle={Proceedings of the IEEE/CVF Conference on Computer Vision and Pattern Recognition},
  pages={11287--11297},
  year={2021}
}

@article{wu2024star,
  title={Star: A benchmark for situated reasoning in real-world videos},
  author={Wu, Bo and Yu, Shoubin and Chen, Zhenfang and Tenenbaum, Joshua B and Gan, Chuang},
  journal={arXiv preprint arXiv:2405.09711},
  year={2024}
}

@article{yi2019clevrer,
  title={Clevrer: Collision events for video representation and reasoning},
  author={Yi, Kexin and Gan, Chuang and Li, Yunzhu and Kohli, Pushmeet and Wu, Jiajun and Torralba, Antonio and Tenenbaum, Joshua B},
  journal={arXiv preprint arXiv:1910.01442},
  year={2019}
}

@inproceedings{yang2021just,
  title={Just ask: Learning to answer questions from millions of narrated videos},
  author={Yang, Antoine and Miech, Antoine and Sivic, Josef and Laptev, Ivan and Schmid, Cordelia},
  booktitle={Proceedings of the IEEE/CVF international conference on computer vision},
  pages={1686--1697},
  year={2021}
}

@article{mahdisoltani2018effectiveness,
  title={On the effectiveness of task granularity for transfer learning},
  author={Mahdisoltani, Farzaneh and Berger, Guillaume and Gharbieh, Waseem and Fleet, David and Memisevic, Roland},
  journal={arXiv preprint arXiv:1804.09235},
  year={2018}
}

@article{li2022uniformerv2,
  title={Uniformerv2: Spatiotemporal learning by arming image vits with video uniformer},
  author={Li, Kunchang and Wang, Yali and He, Yinan and Li, Yizhuo and Wang, Yi and Wang, Limin and Qiao, Yu},
  journal={arXiv preprint arXiv:2211.09552},
  year={2022}
}

@article{liu2024bench,
  title={Et bench: Towards open-ended event-level video-language understanding},
  author={Liu, Ye and Ma, Zongyang and Qi, Zhongang and Wu, Yang and Shan, Ying and Chen, Chang W},
  journal={Advances in Neural Information Processing Systems},
  volume={37},
  pages={32076--32110},
  year={2024}
}

@inproceedings{bain2021frozen,
  title={Frozen in time: A joint video and image encoder for end-to-end retrieval},
  author={Bain, Max and Nagrani, Arsha and Varol, G{\"u}l and Zisserman, Andrew},
  booktitle={Proceedings of the IEEE/CVF international conference on computer vision},
  pages={1728--1738},
  year={2021}
}

@article{wu2025large,
  title={A large cross-modal video retrieval dataset with reading comprehension},
  author={Wu, Weijia and Zhao, Yuzhong and Li, Zhuang and Li, Jiahong and Zhou, Hong and Shou, Mike Zheng and Bai, Xiang},
  journal={Pattern Recognition},
  volume={157},
  pages={110818},
  year={2025},
  publisher={Elsevier}
}

@misc{zhang2024videoinstructiontuningsynthetic,
    title={Video Instruction Tuning With Synthetic Data}, 
    author={Yuanhan Zhang and Jinming Wu and Wei Li and Bo Li and Zejun Ma and Ziwei Liu and Chunyuan Li},
    year={2024},
    eprint={2410.02713},
    archivePrefix={arXiv},
    primaryClass={cs.CV},
    url={https://arxiv.org/abs/2410.02713}, 
}
}

\clearpage
\setcounter{page}{1}
\maketitlesupplementary
\appendix

\section{Training Recipe}
The complete list of training datasets for UFVideo is detailed in \cref{tab:training_recipe}. Utilizing a two-stage training approach, we initially pre-train the model to integrate the tasks of temporal grounding and video segmentation, addressing the limitations of the VideoRefer-7B model \cite{yuan2025videorefer} which lacks these capabilities. We draw upon the temporal training datasets from LLaVA-ST \cite{li2025llava} and Distime \cite{zeng2025distime}, adapting them with relative temporal tokens to better suit our model's requirements. For video segmentation, we use commonly employed video segmentation training datasets to synchronize the LLM output embedding with the mask decoder.
In the second stage, we compile a hybrid large-scale dataset encompassing global video question answering, pixel-level video object referring, video segmentation, temporal video grounding, and self-constructed multi-grained video cooperative understanding tasks. This dataset contains over 3 million samples, and we plan to open source it soon!

\begin{table}[htbp]
\setlength{\tabcolsep}{3mm} 
\centering
\caption{The training recipe of UFVideo, which joint common video understanding tasks such as general video understanding, pixel-level video obejct referring, pixel-level video segmentation and temporal video grounding tasks, and multi-grained video cooperative understanding tasks.}
\label{tab:training_recipe}
\renewcommand{\arraystretch}{1}
\adjustbox{width=1\linewidth}{
   \begin{tabular}{l|lc|cc}
       \toprule[1.2pt]
            \multicolumn{1}{c|}{\textbf{Stage}}
            &\multicolumn{1}{l}{\textbf{Dataset}}
            &\multicolumn{1}{c|}{\textbf{Task}}
            &\multicolumn{1}{c}{\textbf{\#Samples}}
            &\multicolumn{1}{c}{\textbf{\#Ratio}} \\  
\midrule[0.5pt]
 \multirow{20}{*}{1} & MeViS \cite{ding2023mevis} & \multirow{4}{*}{Segmemtation} & 23K & 0.5  \\
 & Ref-DAVIS17 \cite{pont20172017} &  & 0.6K & 0.5  \\
 & Ref-YouTube-VOS \cite{seo2020urvos} &  & 13K & 0.5  \\
 & ReVOS \cite{yan2024visa} &  & 29K & 0.5  \\
 \cmidrule{2-5}
 & Anet-Caption \cite{krishna2017dense} & \multirow{16}{*}{Temporal} & 37K & 0.5  \\
 & Anet-Caption-expand \cite{zeng2025distime} & & 37K & 0.5  \\
 & Anet-RTL \cite{huang2024lita} &  & 10K & 0.5  \\
 & COIN \cite{tang2019coin} &  & 10K & 0.5 \\
 & DiDeMo \cite{anne2017localizing} &  & 33K & 0.5  \\
 & ET-Instruct \cite{liu2024bench} &  & 72K & 0.5  \\
 & Grounded-VideoLLM \cite{wang2024grounded} &  & 17K & 0.5  \\
 & HiREST \cite{zala2023hierarchical} &  & 0.8K & 0.5 \\
 & InterVid-G \cite{wang2403internvideo2} &  & 134K & 0.5  \\
 & InternVid-TG \cite{zeng2025distime} &  & 86K & 0.5  \\
 & Moment-10M \cite{qian2024momentor} &  & 60K & 0.5  \\
 & QuerYD \cite{oncescu2021queryd} &  & 34K & 0.5 \\
 & VTG-IT \cite{guo2025vtg} &  & 32K & 0.5 \\
 & ViTT \cite{huang2020multimodal} &  & 5K & 0.5 \\
 & VTimeLLM \cite{huang2024vtimellm} &  & 139K & 0.5 \\
 & YouCook2 \cite{zhou2018towards} &  & 1.8K & 0.5 \\
 \midrule[0.5pt]
 \multirow{40}{*}{2} & AGQA \cite{grunde2021agqa} & \multirow{17}{*}{General QA} & 1600K & 1  \\
 & CLEVRER \cite{yi2019clevrer} &  & 82K & 1  \\
 & EgoQA \cite{grauman2022ego4d} &  & 21K & 1 \\
 & Intent-QA \cite{li2023intentqa} &  & 12K & 1  \\
 & Kinetics \cite{li2022uniformerv2} &  & 26K & 1 \\
 & LLaVA-Video-178K \cite{zhang2024videoinstructiontuningsynthetic} &  & 300K & 1 \\
 & NExT-QA \cite{xiao2021next} &  & 34K & 1  \\
 & ShareGPT4Video \cite{chen2024sharegpt4video} &  & 39K & 1 \\
 & STAR \cite{wu2024star} &  & 45K & 1 \\
 & SthSthV2 \cite{mahdisoltani2018effectiveness} &  & 40K & 1  \\
 & TextVR \cite{wu2025large} &  & 39K & 1 \\
 & VCG-Plus-112K \cite{maaz2024videogpt+} &  & 112K & 1  \\
 & Videochat2-Conv \cite{li2023videochat} &  & 9K & 1  \\
 & Videochatgpt-100K \cite{maaz2023video} & & 100K & 1  \\
 & WebVid \cite{bain2021frozen} &  & 400K & 1 \\
 & WebVid-QA \cite{yang2021just} &  & 86K & 1  \\
 & YouCook2 \cite{zhou2018towards} &  & 8K & 1 \\
 \cmidrule{2-5}
 & MeViS \cite{ding2023mevis} & \multirow{4}{*}{Segmemtation} & 23K & 5  \\
 & Ref-DAVIS17 \cite{pont20172017} &  & 0.6K & 10  \\
 & Ref-YouTube-VOS \cite{seo2020urvos} &  & 13K & 7  \\
 & ReVOS \cite{yan2024visa} &  & 29K & 5  \\
 \cmidrule{2-5}
 & VideoRefer-700K (Detail) \cite{grunde2021agqa} & \multirow{3}{*}{Referring} & 125K & 2  \\
 & VideoRefer-700K (Short) \cite{grunde2021agqa} &  & 50K & 1  \\
 & VideoRefer-700K (QA) \cite{grunde2021agqa} &  & 75K & 2  \\
 \cmidrule{2-5}
 & Anet-Caption \cite{krishna2017dense} & \multirow{16}{*}{Temporal} & 37K & 1  \\
 & Anet-Caption-expand \cite{zeng2025distime} & & 37K & 1  \\
 & Anet-RTL \cite{huang2024lita} &  & 10K & 1  \\
 & COIN \cite{tang2019coin} &  & 10K & 1 \\
 & DiDeMo \cite{anne2017localizing} &  & 33K & 1  \\
 & ET-Instruct \cite{liu2024bench} &  & 72K & 1  \\
 & Grounded-VideoLLM \cite{wang2024grounded} &  & 17K & 1  \\
 & HiREST \cite{zala2023hierarchical} &  & 0.8K & 1 \\
 & InterVid-G \cite{wang2403internvideo2} &  & 134K & 1  \\
 & InternVid-TG \cite{zeng2025distime} &  & 86K & 1  \\
 & Moment-10M \cite{qian2024momentor} &  & 60K & 1  \\
 & QuerYD \cite{oncescu2021queryd} &  & 34K & 1 \\
 & VTG-IT \cite{guo2025vtg} &  & 32K & 1 \\
 & ViTT \cite{huang2020multimodal} &  & 5K & 1 \\
 & VTimeLLM \cite{huang2024vtimellm} &  & 139K & 1 \\
 & YouCook2 \cite{zhou2018towards} &  & 1.8K & 1 \\
            \bottomrule[1.2pt]
       \end{tabular}
   }
    \vspace{-0.3cm}
\end{table}

\section{UFVideo-Dataset}
\subsection{Construction Details}

The training dataset of PixRQA and PixHQA are constructed using object-level QA data from the VideoRefer-700K dataset \cite{yuan2025videorefer}.The benchmark of PixRQA is built from the VideoRefer-Bench-D dataset and PixHQA is built from the VideoRefer-Bench-Q dataset. The VideoRefer-700K dataset includes masks for referring objects in corresponding frames. For our task definition, in PixRQA, we randomly select a mask from one of the initial frames as the visual prompt, with the remaining frames serving as label masks. In PixHQA, all frame masks are used as label masks. 
Since the frames with masks in the VideoRefer data are not continuous, but PixTRQA requires answering the corresponding time and masks within that range, continuous frames must be maintained. Therefore, the training dataset and benchmark of PixTRQA are based on the ReVOS \cite{yan2024visa} training set and validation set, respectively., which involves segmentation based on an implicit question, ensuring that the object only appears during the relevant time in the video. We also filter the ReVOS data, retaining samples with continuous frames and removing those with dispersed occurrences to ensure a complete time segment as the label.
Thus, all three tasks retain corresponding masks and times, necessitating the reconstruction of semantic questions and responses. We selected the latest Qwen3VL-235B-A22B-Instruct model \cite{qwen3technicalreport} for semantic question-answer construction.

\subsection{UFVideo Dataset Prompt Template}
\label{sec:prompt1}

We design three prompt templates to construct the training datasets for UFVideo's three tasks. The detailed prompts are presented in~\cref{fig:prompt1}.

\subsection{Qwen3-VL Annotation Prompt Template}
\label{sec:prompt2}

To annotate high-quality data, we selected the latest Qwen3-VL-235B-A22B-Instruct as our semantic annotation model, which the prompts shown in~\cref{fig:prompt2}. For the input video and its corresponding object boxes, we generate short descriptions as prompts and long descriptions as semantic responses.

\section{Experiments}

\subsection{Evaluation Metrics}

For general video understanding tasks, as presented in~\cref{tab:mvbench}, the metrics are present in order:
Action Sequence (AS), Action Prediction (AP), Action Antonym (AA), Fine-grained Action (FA), Unexpected Action (UA), Object Existence (OE), Object Interaction (OI), Object Shuffle (OS), Moving Direction (MD), Action Localization (AL), Scene Transition (ST), Action Count (AC), Moving Count (MC), Moving Attribute (MA), State Change (SC), Fine-grained Pose (FP), Character Order (CO), Egocentric Navigation (EN), Episodic Reasoning (ER), Counterfactual Inference (CI). 

For video object referring tasks, we evaluate on VideoRefer-Bench-D and VideoRefer-Bench-Q, where VideoRefer-Bench-D metrics are Subject Correspondence (SC), Appearance Description (AD), Temporal Description (TD) and Hallucination Detection (HD) and use GPT-4o-2024-08-06 to evaluate semantic scores from 0 to 5. VideoRefer-Bench-Q is multi-choice question, the metrics are choice accuracy, including four dimensions of understanding, including Basic Questions (BQ), Sequential Questions (SQ), Relationship Questions (RQ), Reasoning Questions (RQ) and Future Predictions (FP). 

For video segmentation tasks, we use region similarity $\mathcal{J}$, contour accuracy $\mathcal{F}$, and their combination $\mathcal{J}\& \mathcal{F}$ as evaluation metrics. 

For temporal video grounding tasks, the primary metrics are tIoU, R@0.3, R@0.5, and R@0.7, where tIoU measures the IoU between the predicted and ground truth time intervals, and R@k represents the proportion of tIoU values exceeding k. 

For UFVideo-Bench, we adopt $\mathcal{J}$, $\mathcal{F}$, and $\mathcal{J}\& \mathcal{F}$ to measure segmentation performance, and SAvg. represents semantic average scores following VideoRefer-Bench-D evaluation settings. tIoU follows the same definition as in temporal grounding tasks. For PixHQA, we distinguish between temporal point-based and temporal interval-based answers, excluding temporal description metrics for point-based answers following VideoRefer-Bench-D evaluation. For PixTRQA, we only compute segmentation mask metrics for frames within time ranges satisfying the tIoU threshold.

\subsection{Qualitative Results}
\begin{figure*}[!h]
\centering
\includegraphics[width=1\textwidth]{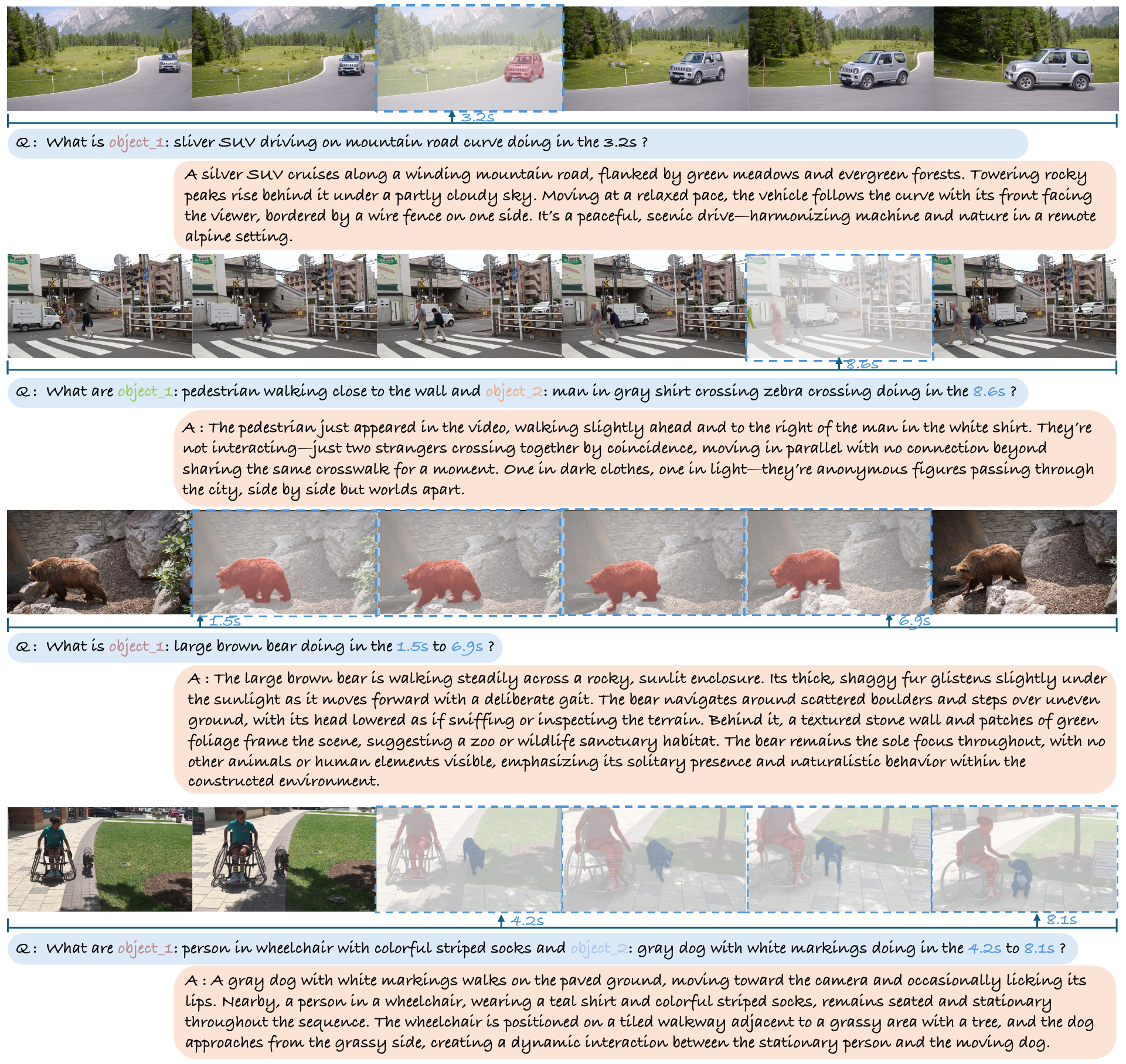}
\caption{Visualized results of PixHQA.}
\vspace{-0.1cm}
\label{fig:pixhqa} 
\end{figure*}

\begin{figure*}[!h]
\centering
\includegraphics[width=1\textwidth]{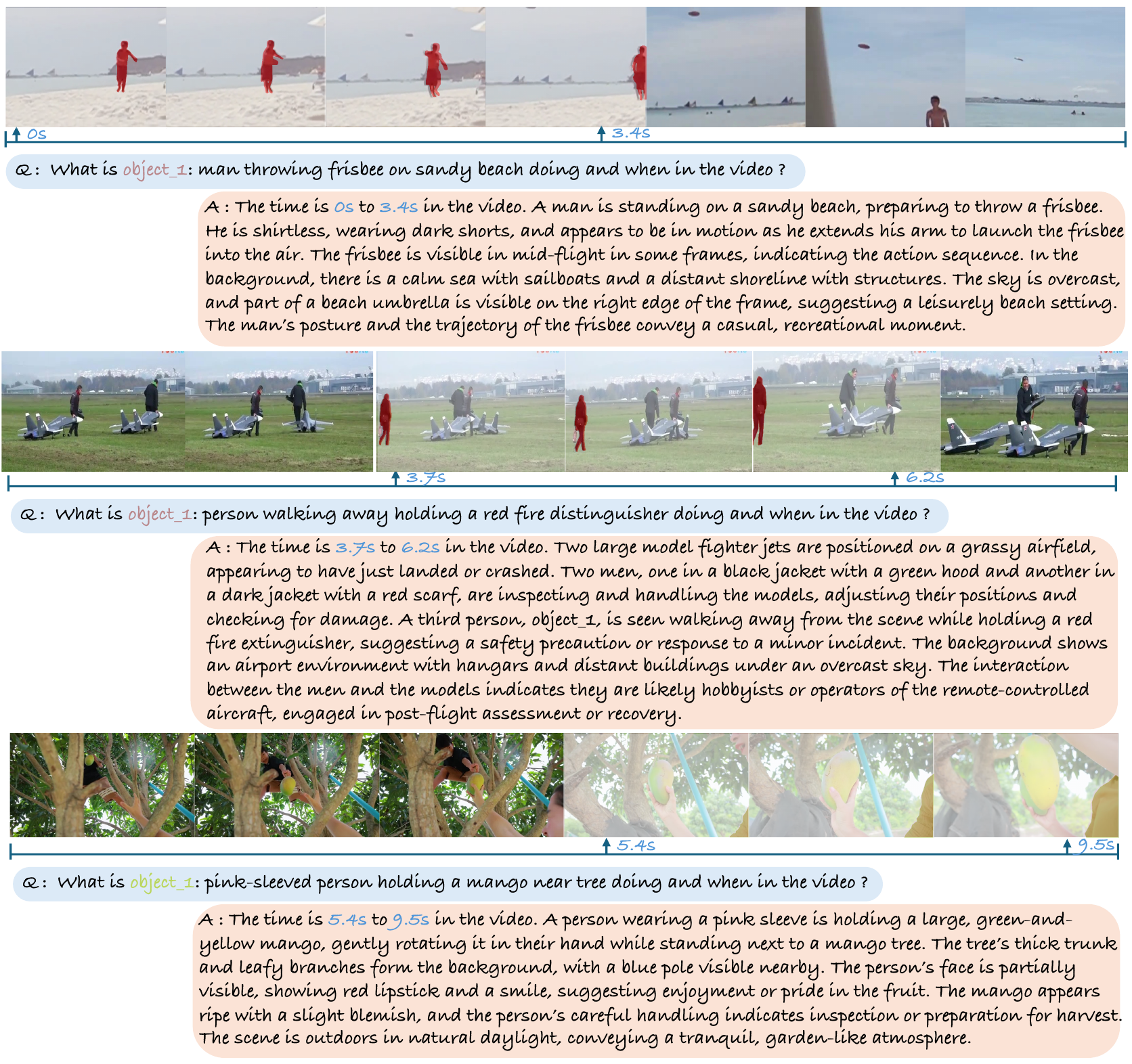}
\caption{Visualized results of PixTRQA.}
\vspace{-0.1cm}
\label{fig:pixtrqa} 
\end{figure*}

\begin{figure*}[!h]
\centering
\vspace{-0.2cm}
\includegraphics[width=1\textwidth]{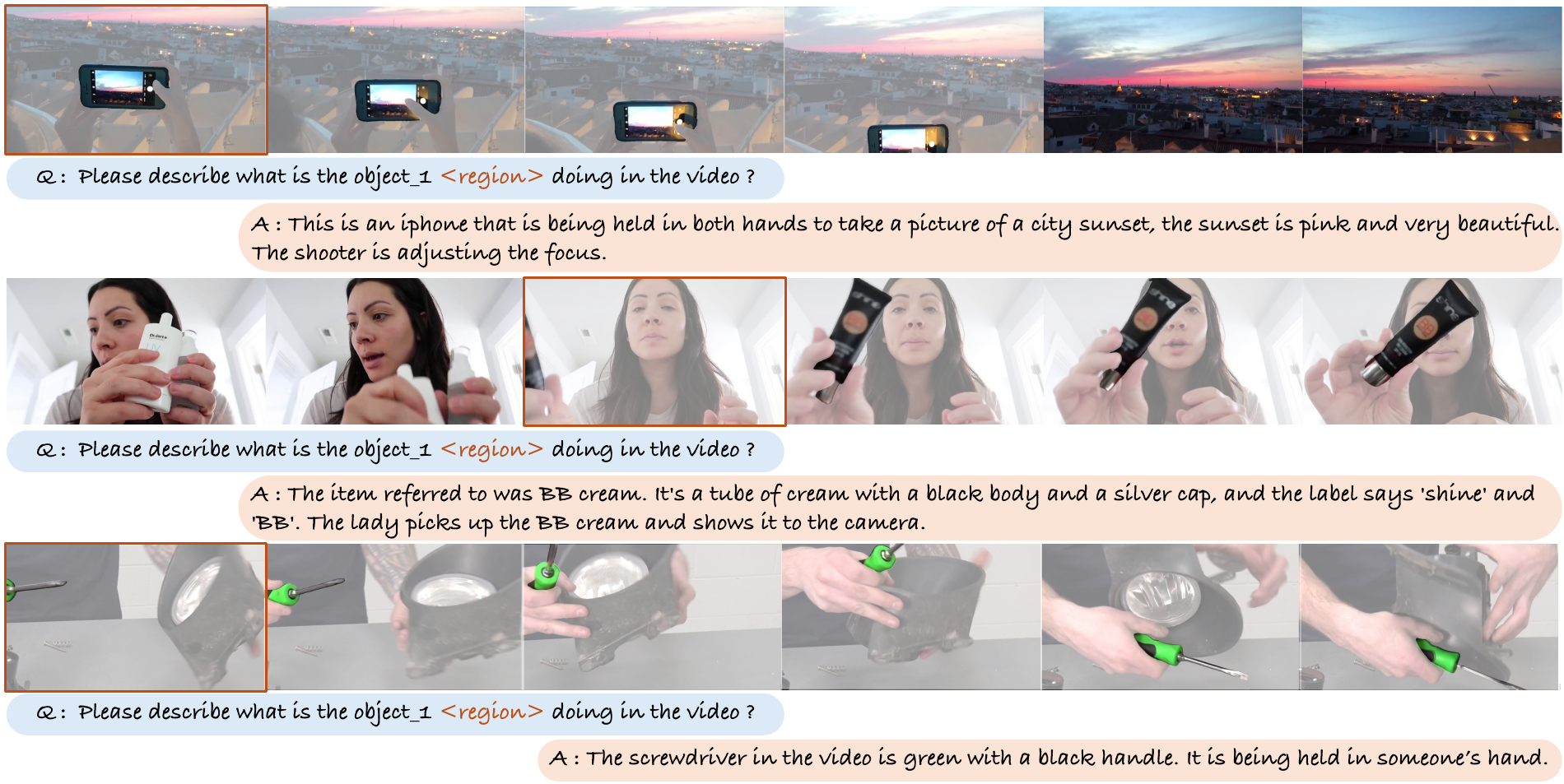}
\caption{Visualized results of PixRQA.}
\vspace{-0.2cm}
\label{fig:pixrqa} 
\end{figure*}

\begin{figure*}[!h]
\centering
\includegraphics[width=1\textwidth]{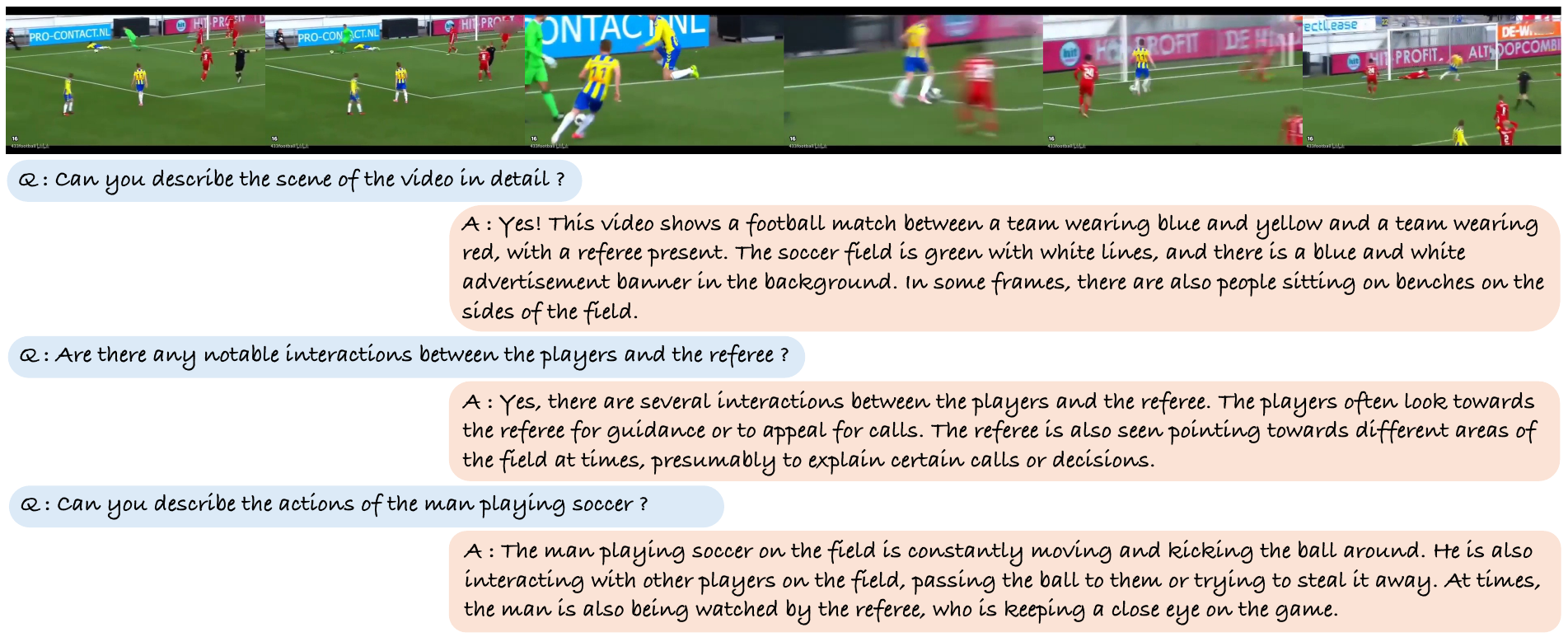}
\caption{Visualized results of multi-turn open-ended video QA of UFVideo.}
\vspace{-0.2cm}
\label{fig:generalQA} 
\end{figure*}

\begin{figure*}[!h]
\centering
\includegraphics[width=1\textwidth]{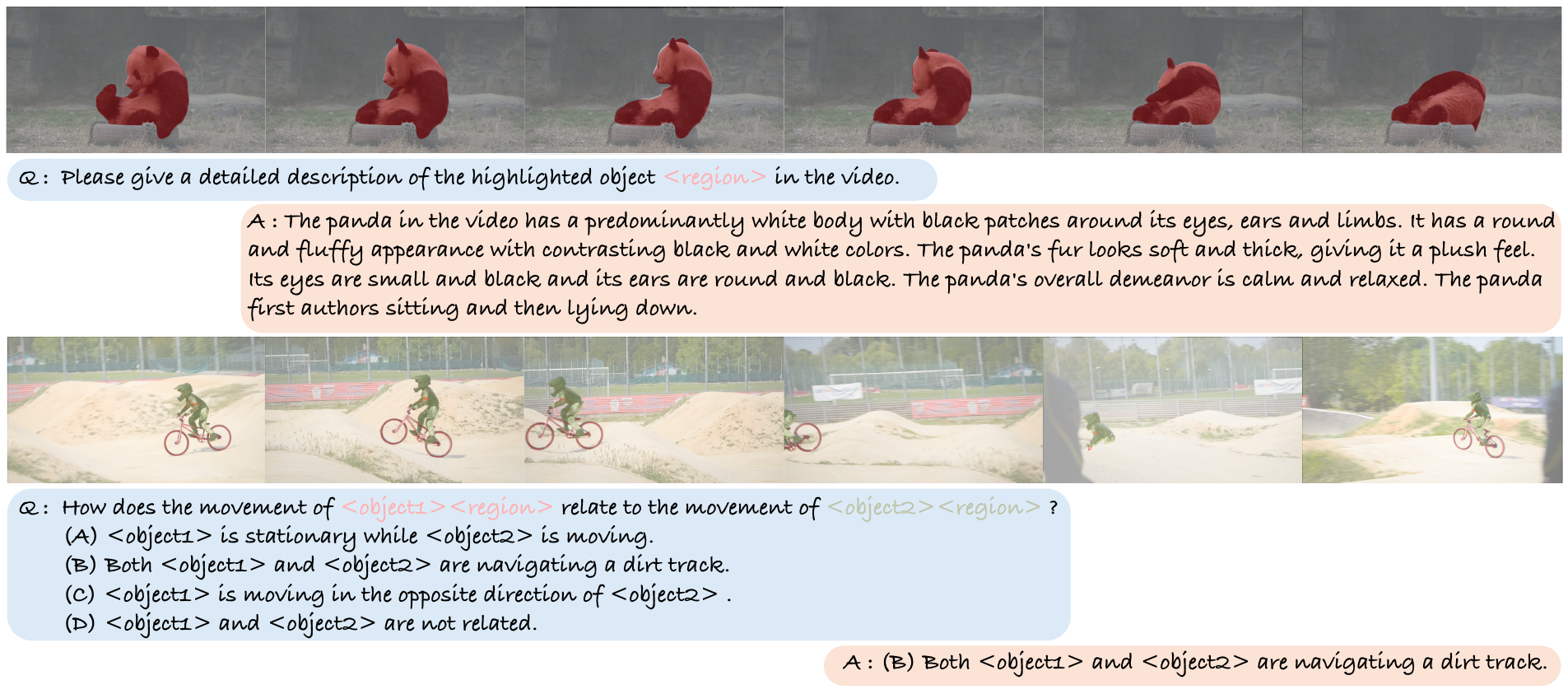}
\caption{Visualized results of VideoRefer-Bench-D (top) and VideoRefer-Bench-Q (bottom).}
\vspace{-0.2cm}
\label{fig:referring} 
\end{figure*}

\begin{figure*}[!h]
\centering
\includegraphics[width=1\textwidth]{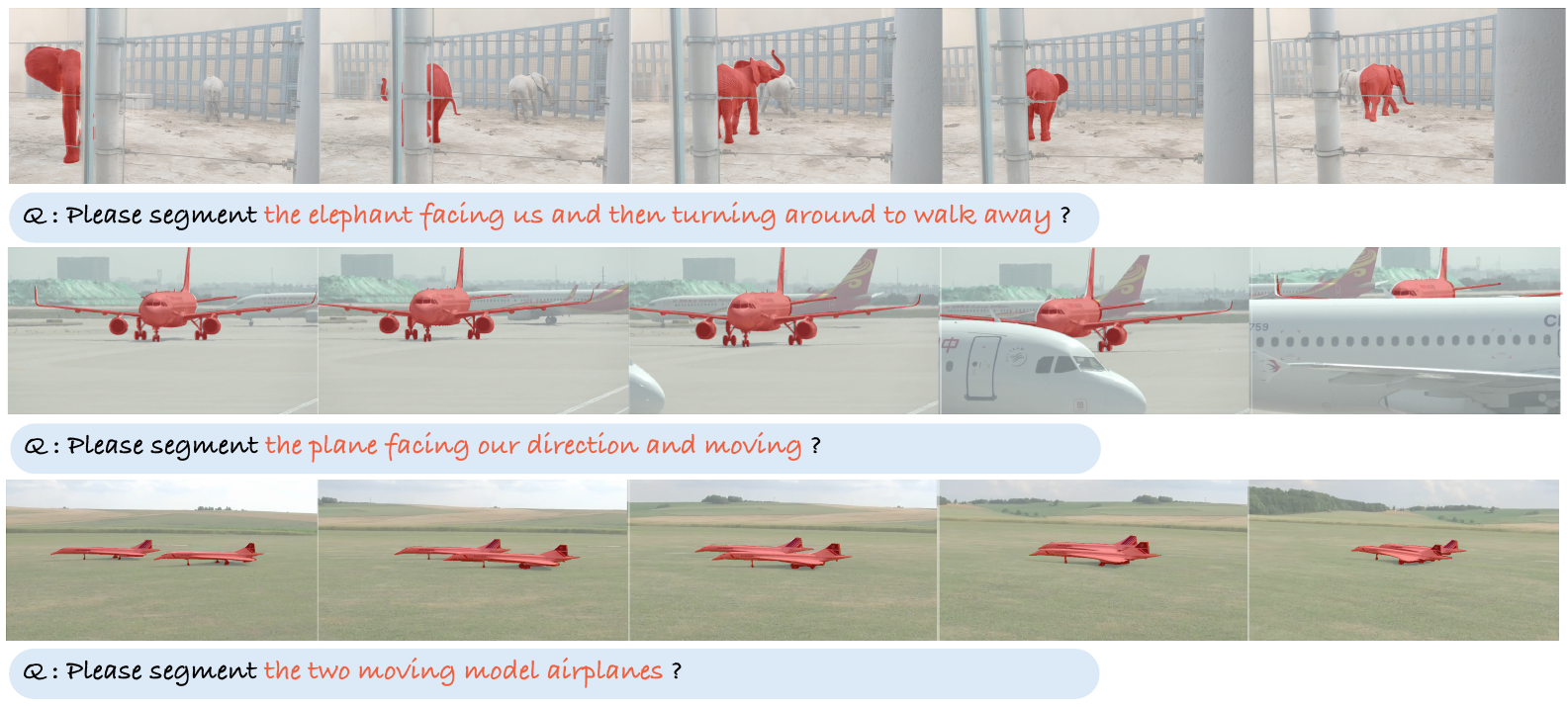}
\caption{Visualized results of MeViS.}
\vspace{-0.2cm}
\label{fig:MeVis} 
\end{figure*}

\begin{figure*}[!h]
\centering
\includegraphics[width=1\textwidth]{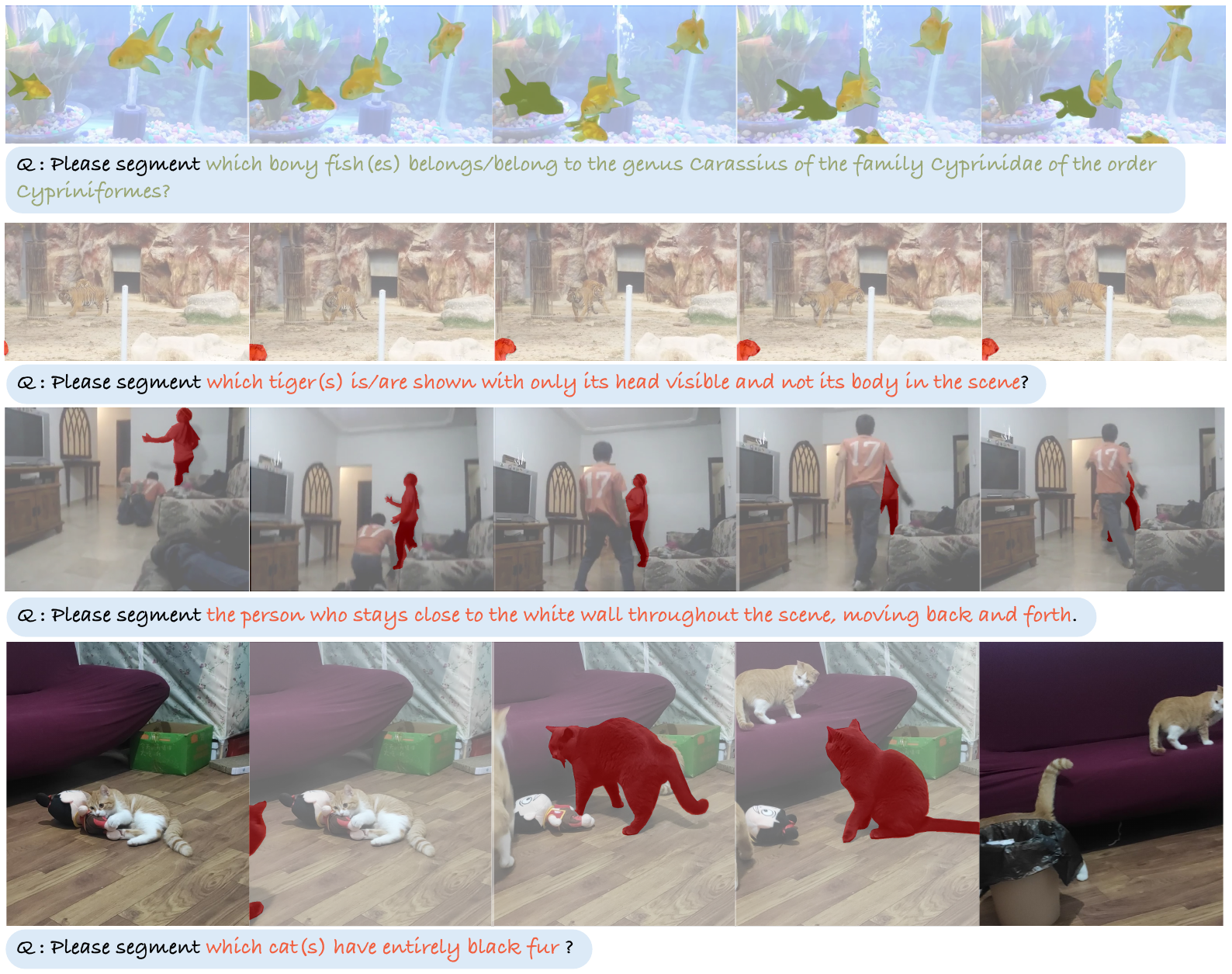}
\caption{Visualized results of ReVOS. The first two rows illustrate the reasoning process of ReVOS, while the last two rows demonstrate the referring aspect of ReVOS.}
\vspace{-0.1cm}
\label{fig:ReVOS} 
\end{figure*}

\begin{figure*}[!h]
\centering
\includegraphics[width=1\textwidth]{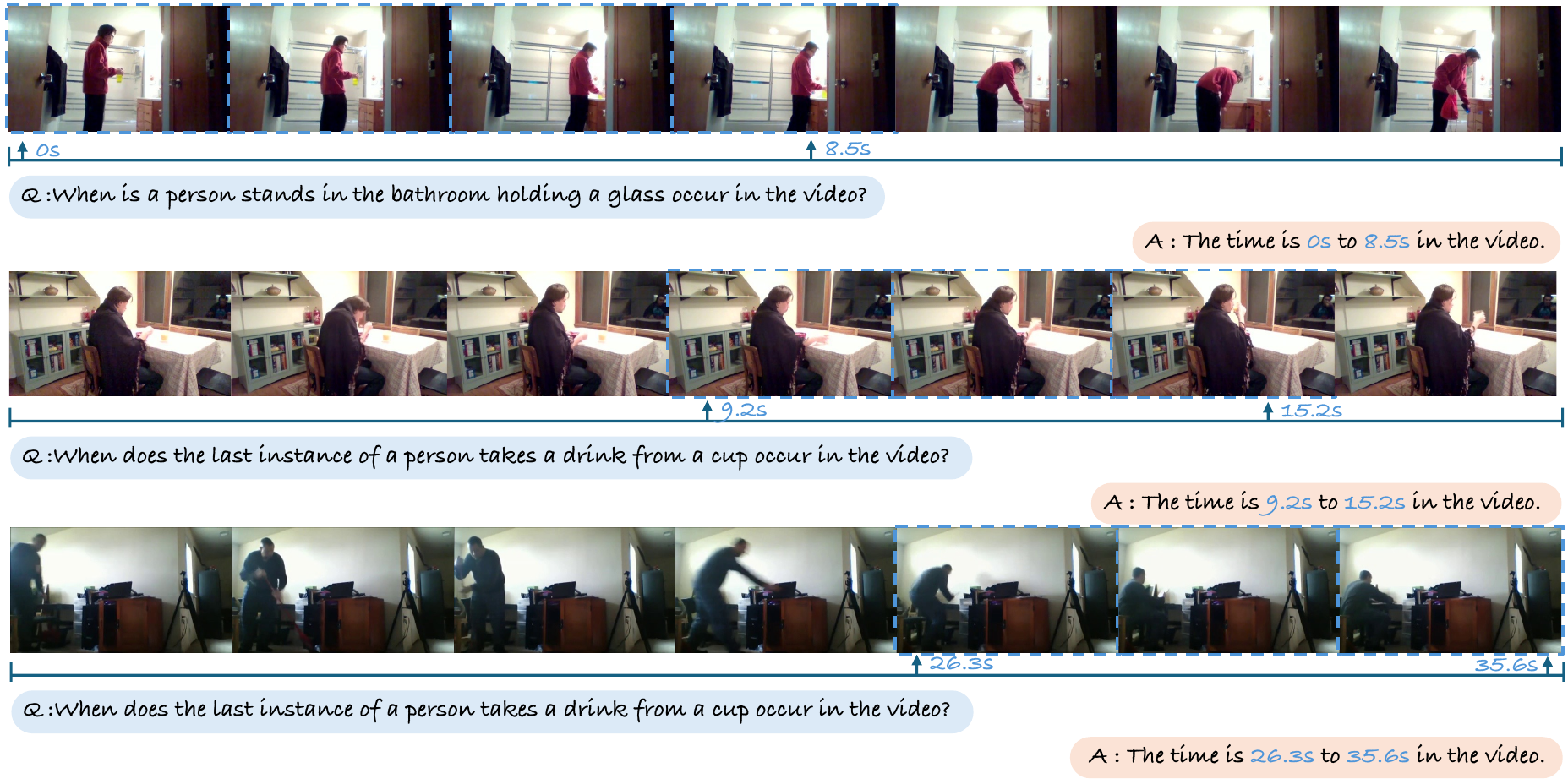}
\caption{Visualized results of Charades-STA.}
\vspace{-0.1cm}
\label{fig:charades} 
\end{figure*}

We present additional outputs from UFVideo for common video understanding tasks and multi-grained video cooperative understanding task. Multi-grained video cooperative understanding results of PixHQA, PixTRQA and PixRQA are presented in~\cref{fig:pixhqa}, ~\cref{fig:pixtrqa} and ~\cref{fig:pixrqa}, respectively. General video understanding results are shown in~\cref{fig:generalQA}, video object referring results in~\cref{fig:referring}, video segmentation results in~\cref{fig:MeVis} and ~\cref{fig:ReVOS}, and temporal video grounding results in~\cref{fig:charades}. These visualizations demonstrate that UFVideo achieves strong performance and flexibly handles multi-grained video understanding tasks.

\subsection{More Experimental Results}

As shown in~\cref{tab:revos}, we present the complete experimental results on the ReVOS dataset for the reasoning video object segmentation task. The results demonstrate that UFVideo outperforms state-of-the-art baselines in both referring and reasoning segmentation, proving that UFVideo gains effective enhancement from multi-grained information.

\begin{table*}[t]
\setlength{\tabcolsep}{12pt} 
\centering
\caption{Experimental results on ReVOS dataset for reasoning video object segmentation task, and comparison with state-of-the-art methods. The red/blue indicates the best/second-best results.}
\renewcommand{\arraystretch}{1}
\adjustbox{width=0.9\linewidth}{
   \begin{tabular}{l|c|ccc|ccc|ccc}
       \toprule[1.2pt]
            \multirow{2}{*}{Model}
            &\multirow{2}{*}{Size}
            &\multicolumn{3}{c|}{Referring}
            &\multicolumn{3}{c|}{Reasoning}
            &\multicolumn{3}{c|}{Overall}\\
\cmidrule{3-5} \cmidrule{6-8} \cmidrule{9-11}  
                & & $\mathcal{J}$ & $\mathcal{F}$ & $\mathcal{J\&F}$
                & $\mathcal{J}$ & $\mathcal{F}$ & $\mathcal{J\&F}$
                & $\mathcal{J}$ & $\mathcal{F}$ & $\mathcal{J\&F}$ \\
\hline
\multicolumn{11}{c}{\cellcolor{color_gray}\textit{Specialist Models}} \\
MTTR \cite{botach2022end} & -- & 29.8 & 30.2 & 30.0 & 20.4 & 21.5 & 21.0 & 25.1 & 25.9 & 25.5  \\
LMPM \cite{ding2023mevis} & -- & 29.0 & 39.1 & 34.1 & 13.3 & 24.3 & 18.8 & 21.2 & 31.7 & 26.4  \\
ReferFormer \cite{wu2022language} & -- & 31.2 & 34.3 & 32.7 & 21.3 & 25.6 & 23.4 & 26.2 & 29.9 & 28.1  \\
\hline
\multicolumn{11}{c}{\cellcolor{color_gray}\textit{Multi-modal LLMs}} \\
LISA \cite{lai2024lisa} & 13B & 45.2 & 47.9 & 46.6 & 34.3 & 39.1 & 36.7 & 39.8 & 43.5 & 41.6 \\
TrackGPT \cite{stroh2024trackgpt} & 13B & 48.3 & 50.6 & 49.5 & 38.1 & 42.9 & 40.5 & 43.2 & 46.8 & 45.0  \\
VISA \cite{yan2024visa} & 13B & 55.6 & 59.1 & 57.4 & 42.0 & 46.7 & 44.3 & 48.8 & 52.9 & 50.9  \\
HyperSeg \cite{wei2024hyperseg} & 3B & 56.0 & 60.9 & 58.5 & 50.2 & 55.8 & 53.0 & 53.1 & 58.4 & 55.7  \\
InstructSeg \cite{wei2024instructseg} & 3B & 54.8 & 59.2 & 57.0 & 49.2 & 54.7 & 51.9 & 52.0 & 56.9 & 54.5 \\
GLUS \cite{lin2025glus} & 7B & 56.0 & 60.7 & 58.3 & 48.8 & 53.9 & 51.4 & 52.4 & 57.3 & 54.9 \\
ViLLa \cite{zheng2025villa} & 6B & -- & -- & -- & -- & -- & -- & 54.9 & 59.1 & 57.0  \\
Sa2VA \cite{yuan2025sa2va} & 4B & -- & -- & -- & -- & -- & -- & -- & -- & 53.2  \\
RGA3 \cite{wang2025object} & 7B & 58.7 & 62.3 & 60.5 & 53.1 & 57.7 & \underline{55.4} & 55.9 & 60.0 & 58.0 \\
UniPixel \cite{liu2025unipixel} & 7B & \underline{64.2} & \underline{68.5} & \underline{66.4} & \underline{59.6} & \textbf{63.9} & \textbf{61.8} & \underline{61.9} & \underline{66.1} & \underline{64.0}  \\
\modelgradient & 7B  & \textbf{65.4} & \textbf{69.8} & \textbf{67.6} & \textbf{59.8} & \underline{63.8} & \textbf{61.8} & \textbf{62.7} & \textbf{66.9} & \textbf{64.8}   \\
            \bottomrule[1.2pt]
       \end{tabular}
   }

   \label{tab:revos}
\end{table*}

\section{Limitation \& Future Work }
In this work, we present UFVideo, which focuses on video-level multi-grained understanding but gives less attention to image-level understanding, which could provide additional enhancement for video comprehension through mutual reinforcement. In future work, we aim to develop a more comprehensive and powerful visual LLM to achieve superior performance on visual cooperative understanding tasks, better serving real-world applications.

\begin{figure*}[t] 
\centering
\begin{tcolorbox}[colback=gray!10, colframe=black, text width=0.9\textwidth, title={Box 1: Prompt Template for PixRQA, PixHQA and PixTRQA tasks.}]
\hypertarget{box4}{}

\textbf{PixRQA: (joint general question answering, video object referring and video segmentation)}

\begin{flushleft}
\textbf{Question}: \texttt{If object\_1 <region> \underline{\texttt{short\, description}}, object\_2 <region> \underline{\texttt{short\, description}}, 
\ldots, object\_n <region> \underline{\texttt{short\, description}}, what is a likely future event? And please generate the mask in every frames.} \\

\textbf{Answer}: \texttt{\underline{\texttt{Long\, description}}. The segmentation mask: object\_1[SEG], object\_2[SEG], \ldots, object\_n[SEG]. } \\
\end{flushleft}

\textbf{PixHQA: (joint general question answering, referring video segmentation and temporal grounding)}

\begin{flushleft}
    
\textbf{Question for timepoint}: \texttt{What object\_1 \underline{\texttt{short\, description}}, object\_2 \underline{\texttt{short\, description}}, \ldots, object\_n \underline{\texttt{short\, description}} are doing in the <Temp-x>, and generate the masks? } \\

\textbf{Question for time period}: \texttt{What object\_1 \underline{\texttt{short\, description}}, object\_2 \underline{\texttt{short\, description}}, \ldots, object\_n \underline{\texttt{short\, description}} are doing in the \{<Temp-x1><Temp-x2>\}, and generate the masks? } \\

\textbf{Answer}: \texttt{\underline{\texttt{Long\, description}}. The segmentation mask: object\_1[SEG], object\_2[SEG], \ldots, object\_n[SEG]. } \\
\end{flushleft}

\textbf{PixTRQA: (joint general question answering, referring video segmentation and temporal retrieval)}
\begin{flushleft}
\textbf{Question}: \texttt{What object\_1 <region> \underline{\texttt{short\, description}}, object\_2 <region> \underline{\texttt{short\, description}}, 
\ldots, object\_n <region> \underline{\texttt{short\, description}} are doing? And please generate the time period and object mask.} \\

\textbf{Answer}: \texttt{The Time is \{<Temp-x1><Temp-x2>\}. \underline{\texttt{Long\, description}}. The segmentation mask: object\_1[SEG], object\_2[SEG], \ldots, object\_n[SEG]. } \\

\end{flushleft}

\end{tcolorbox}
\caption{Prompt template for three UFVideo-Bench tasks. \texttt{<Region>} denotes as the special token for video object referring task. \texttt{[SEG]} denotes as the special token for video segmenatation, \texttt{<Temp-x>} represents the special token for temporal video grounding. }
\label{fig:prompt1}
\end{figure*}

\begin{figure*}[t] 
\centering

\begin{tcolorbox}[colback=gray!10, colframe=black, text width=0.9\textwidth, title={Box 2: Prompt template for annotation in Qwen3-VL-235B-A22B-Instruct}]
\hypertarget{box4}{}

\begin{flushleft}
\textbf{Question} :

\texttt{I upload multiple frames of the video as single-frame images, arranged in the original video order. Next, I upload the box positions of single or multiple objects in each frame as lists. Please provide a $<$Short Description$>$ of about 10 words for each object, and a $<$Long Description$>$ of all objects and their relationships. Both descriptions apply to the entire video, not individual frame. In the $<$Long Description$>$, use the object nouns in the $<$Short Description$>$ to refer to each object. All descriptions should be in English.}  \\

\texttt{Object box lists: {box\_list\_prompt}.}

\texttt{Provide your description strictly in the following format:   
$<$Short Description$>$:\{object short description prompt\}   
$<$Long Description$>$: your long description of all objects and their relationships.} \\

\textbf{Answer Examples}:

\underline{Single object answer}:

\texttt{$<$Short Description$>$:
object\_1: Blue monster truck mid-air during stunt jump}

\texttt{$<$Long Description$>$:
The blue monster truck, object\_1, is captured mid-air, performing a high jump over a dirt track. In the background, several other monster trucks are visible on the ground, including a white one and additional blue ones, positioned along the track as if awaiting their turn or observing the stunt. The scene takes place in an outdoor arena with spectators behind barriers and trees and buildings in the distance, suggesting a live event or competition. The blue monster truck’s shadow is cast on the ground below, emphasizing its height and motion.}

\underline{Multi objects answer}:

\texttt{$<$Short Description$>$:
object\_1: Blue monster truck soaring through the air.
object\_2: Line of monster trucks on dirt track.}

\texttt{$<$Long Description$>$:
The blue monster truck, object\_1, is captured mid-air, performing a high jump over the dirt track. Below it, object\_2, a line of monster trucks, is parked or moving slowly along the track, creating a dynamic contrast between the airborne vehicle and the grounded ones. The scene unfolds at an outdoor event with spectators visible in the background, emphasizing the spectacle of the stunt.}

\end{flushleft}

\end{tcolorbox}
\caption{Prompt template for annotation in Qwen3-VL-235B-A22B-Instruct. We present the question template and answer examples with single or multi objects.}
\label{fig:prompt2}
\end{figure*}

\end{document}